\documentclass[10pt,journal,compsoc]{IEEEtran}
\ifCLASSOPTIONcompsoc
  \usepackage[nocompress]{cite}
\else
  \usepackage{cite}
\fi
\ifCLASSINFOpdf
\else
\fi
\usepackage{graphicx}
\usepackage{times}
\usepackage{epsfig}
\usepackage{amsmath}
\usepackage{bm}
\usepackage{amssymb}
\usepackage{ltablex}
\usepackage{stmaryrd}
\usepackage{mathtools}
\usepackage{multicol}
\usepackage{multirow}
\usepackage{lipsum}
\usepackage{bbm}
\usepackage{soul}
\usepackage{color}
\usepackage{adjustbox}

\usepackage[linesnumbered,ruled,vlined]{algorithm2e}

\SetCommentSty{mycommfont}
\SetKwInput{KwData}{Input}
\SetKwInput{KwResult}{Output}

\newcommand{\RNum}[1]{\uppercase\expandafter{\romannumeral #1\relax}}

\newcommand{\best}[1]{{\bf{#1}}}

\newcommand{\zht}[1]{{\color{black} {#1}}}

\DeclareMathOperator{\VSE}{VSE}

\DeclareMathOperator{\tensor}{tensor}

\DeclareMathOperator{\convd}{conv3d}
\DeclareMathOperator{\view}{view}
\DeclareMathOperator{\cat}{cat}
\DeclareMathOperator{\gridsample}{gridsample}
\DeclareMathOperator{\sumtorch}{sum}
\DeclareMathOperator{\unsqueeze}{unsqueeze}
\DeclareMathOperator{\aaa}{expand}

\DeclareMathOperator{\maxtorch}{max}
\DeclareMathOperator{\gathertorch}{gather}
\DeclareMathOperator{\clamp}{clamp}

\DeclareMathOperator*{\argmax}{arg\,max}

\DeclarePairedDelimiter{\norm}{\lVert}{\rVert} 
\begin{document}
%
\title{Semantic Layout Manipulation with High-resolution Sparse Attention}

\author{{Haitian Zheng$^{1*}$} \  Zhe Lin$^{2}$ \  Jingwan Lu$^{2}$ \  Scott Cohen$^{2}$ \  Jianming Zhang$^{2}$ \  Ning Xu$^{2}$ \  Jiebo Luo$^{1}$\\
{$^{1}$}University of Rochester \quad {$^{2}$}Adobe Research\\
$^{1}${\tt\small \{hzheng15,jluo\}@cs.rochester.edu} \quad $^{2}${\tt\small \{zlin,jlu,scohen,jianmzha,nxu\}@adobe.com}
}

\author{
        Haitian~Zheng,~\IEEEmembership{Student Member,~IEEE,}
        Zhe~Lin,~\IEEEmembership{Member,~IEEE,}
        Jingwan~Lu,~\IEEEmembership{Member,~IEEE,}
        Scott~Cohen,~\IEEEmembership{Member,~IEEE,}
        Jianming~Zhang,~\IEEEmembership{Member,~IEEE,}
        Ning~Xu,~\IEEEmembership{Member,~IEEE,}
        Jiebo~Luo,~\IEEEmembership{Fellow,~IEEE,}
\IEEEcompsocitemizethanks{\IEEEcompsocthanksitem Haitian~Zheng and Jiebo~Luo are with the Department
of Computer Science, University of Rochester.\protect\\
E-mail: \{hzheng15,jluo\}@cs.rochester.edu
\IEEEcompsocthanksitem Zhe~Lin, Jingwan~Lu, Scott~Cohen, Jianming~Zhang and Ning~Xu are with Adobe Research..\protect\\
E-mail: \{zlin,jlu,scohen,jianmzha,nxu\}@adobe.com}
}

\markboth{}%
{Shell \MakeLowercase{\textit{et al.}}: Bare Demo of IEEEtran.cls for Computer Society Journals}
%



\IEEEtitleabstractindextext{%
\begin{abstract}
We tackle the problem of semantic image layout manipulation, which aims to manipulate an input image by editing its semantic label map. A core problem of this task is how to transfer visual details from the input images to the new semantic layout while making the resulting image visually realistic. Recent work on learning cross-domain correspondence has shown promising results for global layout transfer with dense attention-based warping. However, this method tends to lose texture details due to the resolution limitation and the lack of smoothness constraint on  correspondence. To adapt this paradigm for the layout manipulation task, we propose a high-resolution sparse attention module that effectively transfers visual details to new layouts at a resolution up to 512x512. To further improve visual quality, we introduce a novel generator architecture consisting of a semantic encoder and a two-stage decoder for coarse-to-fine synthesis. Experiments on the ADE20k and Places365 datasets demonstrate that our proposed approach achieves substantial improvements over the existing inpainting and layout manipulation methods.
\end{abstract}

\begin{IEEEkeywords}
Image manipulation and editing, image synthesis, correspondence learning, inpainting.
\end{IEEEkeywords}}

\maketitle

\IEEEdisplaynontitleabstractindextext

%
\IEEEpeerreviewmaketitle

\begin{figure}[t]
	\centering
	\includegraphics[width=1.\linewidth]{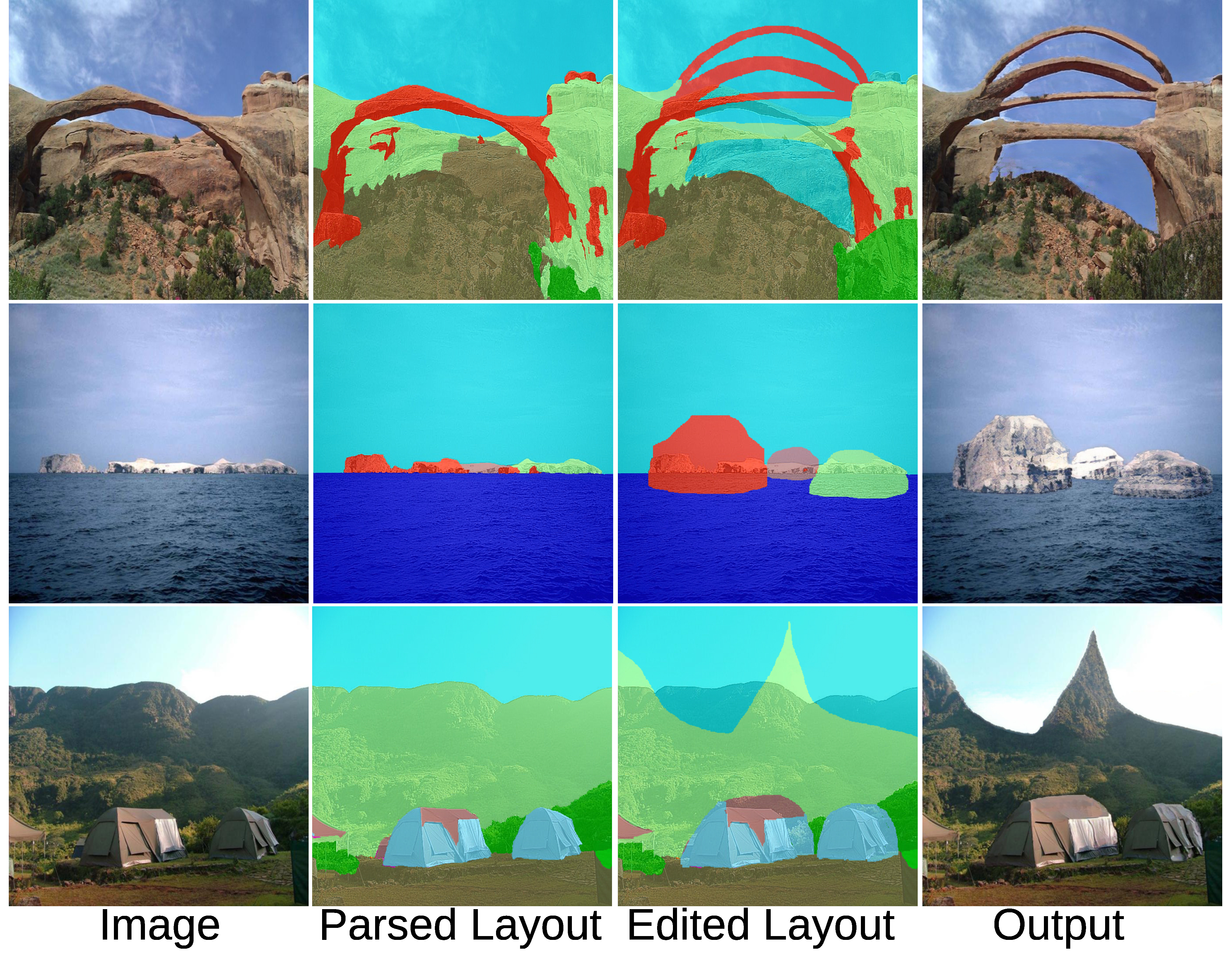}
	\caption{\textbf{Semantic layout manipulation.} Given an input image (1st column) and a semantic label map (3rd column) manipulated from an existing semantic map (2nd column), our network generates manipulated images (last column) that conform to the semantic layout guidance. The images are generated at resolution $512\times 512$.
	It is worth noting that rather than relying on the ground-truth layout for manipulation, our model utilizes a semantic parser~\cite{hrnet} to generate layouts from input images.
	}
	\label{fig:teaser}
\end{figure}

\IEEEraisesectionheading{\section{Introduction}\label{sec:intro}}
Semantic layout manipulation refers to the task of editing an image by modifying its semantic label map, i.e., changing the semantic layout or inserting/erasing objects as illustrated in Fig.~\ref{fig:teaser}. It has many practical image editing applications such as photo-editing~\cite{bau_semantic}, image retargeting~\cite{patchmatch}, restoration~\cite{sesame}, composition~\cite{SemanticPyramid} and image melding~\cite{image_melding}, but is relatively under-explored due to the challenges of predicting complex, non-rigid spatial deformations and the domain gap between the input image and the target semantic layout. Essentially, developing an effective method to transfer visual patterns from the input image to the target semantic layout is the key to solving the problem.

Early works~\cite{image_analogies,patchmatch,cho2008patch,Image_quilting} on image synthesis allow users to mark semantic regions for guided image manipulation. These methods utilize non-parametric modeling of patches~\cite{image_analogies,Image_quilting} or patch-based copy-pasting strategies~\cite{patchmatch,cho2008patch,image_melding,simakov2008summarizing,barnes2010generalized} to generate new images. They work well for generating stationary textures or repeating structures, but cannot hallucinate new semantic structures.

Recently, with the development of deep generative models such as Generative Adversarial Networks (GANs), structure-guided image manipulation has made remarkable progresses. In particular, Champandard~\textit{et al.}~\cite{champandard2016semantic} design a semantic-aware loss on a pretrained classifier~\cite{vgg} for semantic-guided artwork generation. Zhu~\textit{et al.}~\cite{zhu2016generative} optimize the latent code of generative model for layout-constrained image generation.
More recently, guided inpainting methods~\cite{deepfillv2,spg,edgeconnect} are proposed to manipulate nature images. Specifically, guided inpainting approaches mask out some regions from images and hallucinate new pixels to encourage the inpainting results to follow the edge scribbles provided by users.  To achieve semantic layout manipulation, a recent work~\cite{sesame} extends inpainting networks by including semantic label maps in the missing region as inputs. Despite showing promising results,~\cite{sesame} is inherently limited in that 1) it discards pixels inside the mask region which may contain important visual details useful for manipulation tasks, and 2) it lacks the spatial alignment mechanism to handle drastic layout change.   

On the other hand, methods utilizing attention-based warping~\cite{cocosnet,msca} have shown promising results for global reference-based layout editing.
Specifically, Zhang~\textit{et al.}~\cite{cocosnet} introduce a cross-domain correspondence network to warp input images to desired layouts.
Zheng~\textit{et al.}~\cite{msca} propose a spatial-channel attention method to align multi-scale visual features to a new layout.
Due to explicit warping, they can handle dramatic layout changes.
However, the global layout editing task does not take into account the contextual information from the input image and manipulation intention from users. Naively applying global warping for local manipulation 
tasks will also modify the surrounding image contents and is undesirable.
More critically, the warping schemes of~\cite{cocosnet,msca} are computationally costly, and thus are limited to low-resolution warping~\cite{cocosnet} or are applied to individual regions independently~\cite{msca}, causing the loss of visual details in the warped image.

Motivated by the above observations, we extend the guided image inpainting framework to also incorporate a warped version of the original image itself as an additional information source to help the local manipulation task. This strategy effectively overcomes the limitations of both guided inpainting and global attention-based warping method for local manipulation tasks:  1) a guided inpainting framework allows it to exploit contextual information sufficiently for smooth generation, and 2) attention warping schemes provide essential texture/structure information from the entire original image including editing regions. To address the resolution limitation of attention-based warping, inspired by the PatchMatch~\cite{patchmatch} algorithm, we design a new algorithm to efficiently compute dense correspondences at high-resolution using sparse sampling of feature matches and iterative match propagation between the original image and edited layout. \zht{Compared to the existing work~\cite{cocosnet} that computes correspondence on the $4\times$ down-sampled images, our designed module can transfer visual details at the full resolution of input images.}

\zht{
To this end, we first introduce a sparse attention module to efficiently warp high-resolution contents from an input image to a target layout. The attention module is based on two key components, namely 1) key index sampling and 2) sparse attentive warping to respectively generate and transfer visual details from a sparse set of features from the original image to each spatial location of the target layout. Next, we propose a unified semantic layout generator network to produce final manipulation results based on the masked original image, target layout, and the warped proxy output. The network is composed of a semantic encoder and a coarse-to-fine synthesis network and is trained based on a novel self-supervised training scheme. \zht{Recently, a concurrent work~\cite{zhou2020full} proposes a recurrent model to compute full-resolution correspondences. Different from~\cite{zhou2020full}, we aim to improve the efficiency of attention. Using our proposed sparse attention module, correspondence can be directly estimated on high-resolution feature maps.
Different from global layout transfer, local manipulation also requires new design and supervision to take into account contextual clues and the manipulation intention.
To address those challenges, we propose a new network architecture and a self-supervised scheme to tackle local editing.}

}


Our main contributions are four-fold:
\begin{itemize}
    \item An efficient sparse attention module to transfer visual details at a high $512\times512$ resolution. 
    \item A unified semantic layout manipulation framework that takes advantages of both guided inpainting and global layout warping techniques. 
    \item A novel self-supervised training scheme to train the semantic layout manipulation network.
    \item Experiments on ADE20k and Places365 showing that our method achieves substantial improvements over the state-of-the-art inpainting and semantic layout manipulation methods.
\end{itemize}

\begin{figure*}[t]
	\centering
	\includegraphics[width=0.8\linewidth]{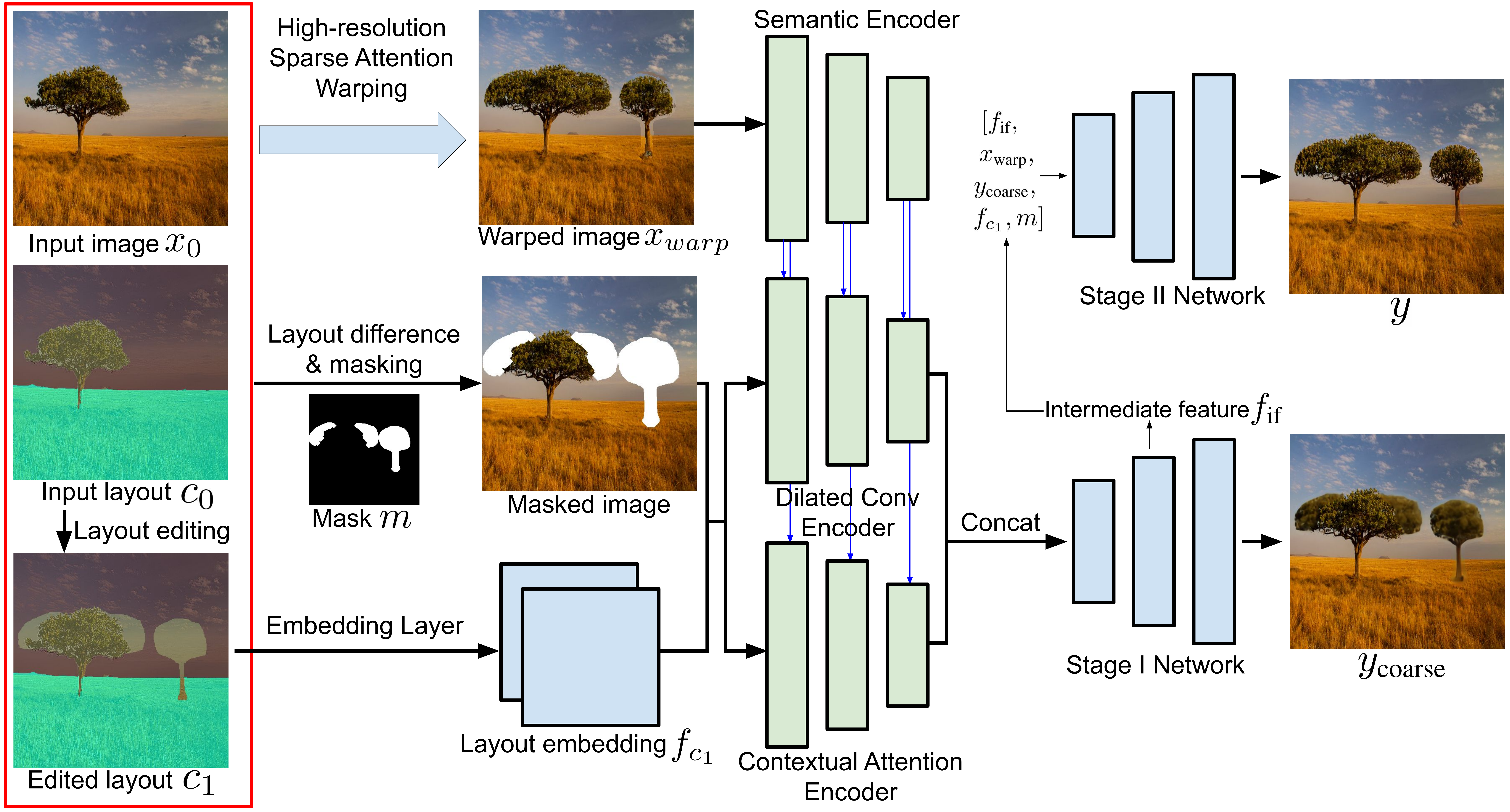}
	\caption{
	Overall pipeline of our semantic layout manipulation framework. In Sec.~\ref{sec:HR_attention}, we first propose a high-resolution sparse attention module (\textbf{left}) to generate the initialed warped image with semantically aligned appearances. Then, in Sec.~\ref{sec:network}, a unified semantic layout generator (\textbf{right}) consisting of a semantic encoder and a coarse-to-fine decoder is proposed to refine the warped image and produce the final result.
	}
	\label{fig:pipeline}
\end{figure*}

\section{Related Work}\label{sec:related}
\noindent \textbf{Image Synthesis and Manipulation.} \quad
Image synthesis and manipulation has a long standing history. Early works on image manipulation apply non-parametric modeling of patches~\cite{efros1999texture,image_analogies} or copy-pasting patches~\cite{patchmatch,cho2008patch,image_melding} to synthesize images. Recently, with the advances of generative adversarial networks (GANs)~\cite{gan,wasserstein,lsgan,improved,sagan,largescalegan,stylegan,progressivegan}, image synthesis and manipulation tasks including style transfer \cite{cyclegan,liu2017unsupervised,munit,drit,pairedcyclegan}, photo-realistic semantic image synthesis~\cite{pix2pix,pix2pixhd,CRN,spade,tan2021efficient,dundar2020panoptic,tan2021diverse}, inpainting~\cite{deepfillv1,deepfillv2,structureflow,spg,edgeconnect} and editing~\cite{zhu2016generative,bau_semantic,liu2020open} have made substantial progresses.

\noindent \textbf{Layout-guided Image Manipulation.} \quad
Layout-guided image manipulation aims to modify an input image such that the generated image conforms to a desired layout.
For layout-based image manipulation, Zhu~\textit{et al.}~\cite{zhu2016generative} apply optimization with layout constraints to update images.
Recently, layout guidance has been used to improve inpainting. In particular, two-stage models~\cite{xiong2019foreground,edgeconnect} are designed to generate edge maps from inputs for better result quality. Song~\textit{et al.}~\cite{spg} use a pre-trained segmentation network to improve inpainting results. Although those works are  designed for inpainting, the guided synthesis framework has inspired a later work~\cite{sesame} for semantic layout manipulation. Specifically, the model of~\cite{sesame} uses local semantic label map as the guidance to manipulate masked images. 

\noindent \textbf{Example-guided Synthesis.} \quad
Example-guided synthesis~\cite{example_cvpr19,cocosnet,msca} aims to transfer the style of an example image to a target condition, (e.g. a semantic label map).
Different from image manipulation, the generated layout can be very different from the exemplar.
Early works leverage image analogies~\cite{image_analogies} and image melding~\cite{image_melding} to transfer the style of an image to a new layout.
To better transfer visual style from the exemplar, neural patch-matching and attention-based warping schemes are later proposed. Specifically, Liao~\textit{et al.} leverage neural feature and bidirectional patchmatch~\cite{simakov2008summarizing} to generate semantic correspondence for visual attribute transfer. Zheng~\textit{et al.}~\cite{msca} propose a decomposed attention scheme for efficient multi-scale feature alignment. Zhang~\textit{et al.}~\cite{cocosnet} propose a cross-domain attention module to transfer pixel from exemplar.

\noindent \textbf{Efficient Attention.} \quad
The attention mechanism~\cite{bahdanau2014neural} is shown effective for layout alignment tasks~\cite{cocosnet,msca,ren2020deep}.
However, they cannot be easily adapted to higher resolution due to the high computational cost.
Recent works focus on approximating the attention with more efficient formulation. Specifically, low-rank decomposition~\cite{doubleattention,msca}, criss-cross attention~\cite{CCA} and Taylor series expansion~\cite{GCNL} are proposed to approximate attention.
In this work, we exploit the spatial coherence of correspondences and sparse sampling to accelerate attention.


\section{Methodology}\label{sec:methodology}
Given an image, an edited semantic layout, and a mask of the region to be modified, our task is to synthesize new image content inside the masked region such that the final completed image is visually plausible and consistent with the edited layout. Existing approaches either apply warping~\cite{cocosnet,msca} or inpainting to generate coherent contents~\cite{sesame}.
However, they lack the merits of their complimentary counterpart, i.e.,~\cite{cocosnet,msca} cannot harmonize the outputs with neighboring regions while~\cite{sesame} lacks spatial alignment and discards pixels inside the mask.

To combine the advantages of both models, we propose a unified generator framework that follows the pipeline of alignment and refinement. Specifically, the alignment stage serves to generate an initial warping that is aligned to the target layout. Next, a refinement network that consists of an encoder and coarse-to-fine decoders serves to refine and harmonize the warped output. Fig.~\ref{fig:pipeline} shows an overview of our manipulation framework. We propose the high-resolution sparse attention (HRS-Att) module to efficiently and accurately transfer high-resolution visual details from the input image to the target label map by attention-based warping. We then propose a two-stage semantic layout manipulation network to generate the final manipulated output by inpainting the masked region with the guidance of the warped output. 


\subsection{Cross-domain Correspondence}
Computing a dense correspondence between the image and the edited layout is a crucial first step for local layout editing, as we need to warp input visual details to fit the target layout while making the warped image visually plausible.
Recently, Zhang et.~al~\cite{cocosnet} propose a cross-domain correspondence network for layout-guided alignment.
\zht{Specifically, it first transforms the input image $x \in \mathbb{R}^{H \times W \times 3}$ and the edited label map $c \in \mathbb{R}^{H \times W \times C}$ (consisting of $C$ semantic classes) to downsampled feature maps that are in a shared domain at resolution
$H/4 \times W/4$:}
\begin{align}
\label{eq:feature_extraction}
\begin{aligned}
    f_x = F_x(x) \quad\text{and}\quad  f_c = F_c(c),
\end{aligned}
\end{align}
where $F_x$ and $F_c$ denotes the cross-domain feature extractor. Here, the output features $f_x,f_c$ are normalized at each location to have zero mean and unit $\ell_2$ norm.
Next, a low-resolution warped image $r^L \in \mathbb{R}^{{H/4 \times W/4 \times 3}}$ is calculated from the downsampled input $x^L \in \mathbb{R}^{{H/4 \times W/4 \times 3}}$ where each pixel $r^L(q)$ is computed as weighted sum of all pixels from $x^L$ \footnote{We use superscript $L$ here to denote low-resolution variables.}.
\begin{align}
\label{eq:warping_dense}
\begin{aligned}
    \displaystyle
    r^L(q) &= \sum_{p} {a}^L_{q p} x^L(p),
\end{aligned}
\end{align}
where $p$ iterates over all spatial locations of $x^L$, ${a}^L_{q p} = {e^{\gamma s^L_{q p}}}/{\sum_{p} {e^{\gamma s^L_{q p}}}}$ is the linear weight computed from the inner-product feature similarity $s^L_{q p}=\langle f_c(q),f_x(p) \rangle$ and $\gamma$ is a temperature coefficient.

\begin{figure}[]
	\centering
	\includegraphics[width=.9\linewidth]{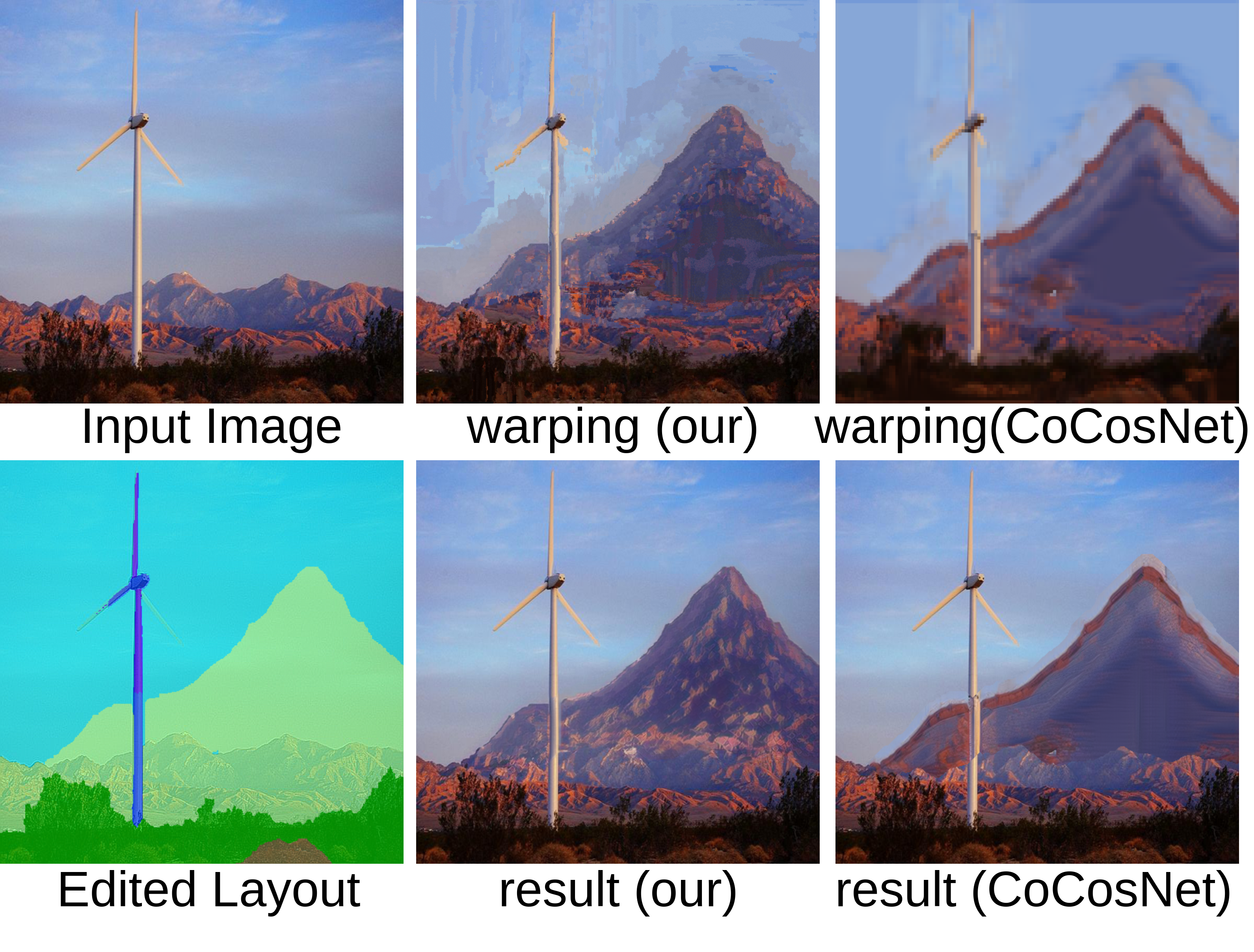}
	\caption{The warping and synthesis results of our method against CoCosNet~\cite{cocosnet}. Our HRS-Att can transfer high-resolution details for better semantic layout manipulation.}
	\label{fig:teaser2}
\end{figure}

The warping function of Eq.~\ref{eq:warping_dense} can be interpreted as an attention operation~\cite{vaswani2017attention}, 
where each index $q$ (query) is mapped to a weighted sum of pixels (value) of index $p$ (key). 
However, such a mapping step needs to compute pair-wise similarity for every query and key, resulting in a quadratic computation complexity w.r.t. the spatial resolution.
Therefore, attention-based correspondence~\cite{cocosnet}
often struggles to transfer high-resolution visual details from the input image as shown in Fig.~\ref{fig:teaser2}.
Note that the detailed texture in the mountains is completely lost in the result of CoCosNet.

\begin{figure}[t]
	\centering
	\includegraphics[width=1.\linewidth]{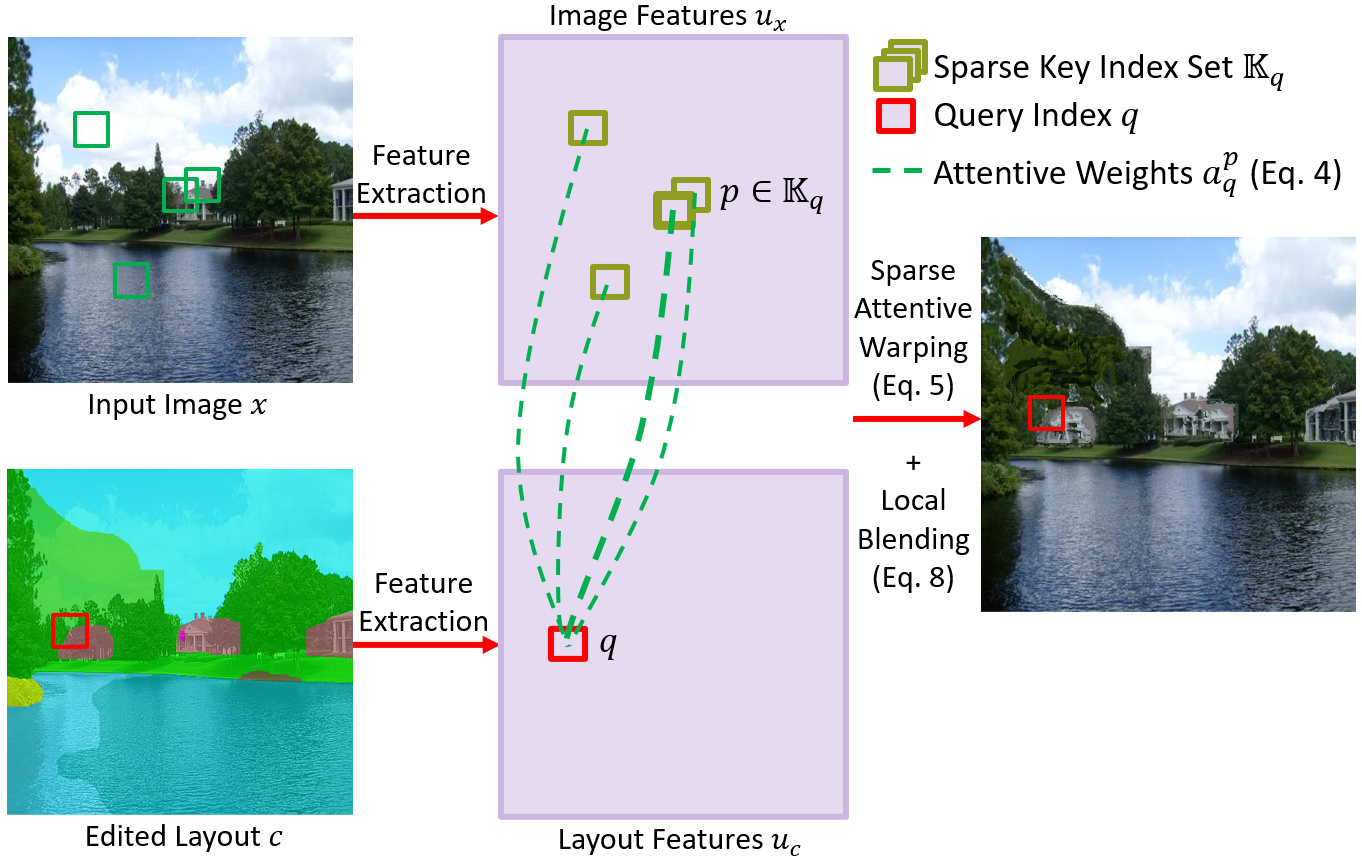}
	\caption{
	The \textbf{H}igh-\textbf{r}esolution \textbf{S}parse \textbf{Att}ention Module (\textbf{HRS-Att}) leverages \textbf{cross-domain feature extractors} to extract high-resolution features from input image and an edited layout. Next, for each query point $q$ from the edited layout, a \textbf{key index sampling step} serves to sample sparse key indexes $p$ for sparse attention computation. Finally, a \textbf{sparse attentive warping} step is proposed to aggregate and warp the input image. More details are described in Sec.~\ref{sec:HR_attention}.
	}
	\label{fig:sparse_attention}
\end{figure}

\subsection{High-resolution Sparse Attention}
\label{sec:HR_attention}
\zht{
Image correspondence is often sparse~\cite{lowe1999object,lucas1981iterative}, meaning that for an image pair $A,B$, a pixel $A$ often matches only with a sparse set of pixels from $B$. However, the vanilla attention is inherently dense and inefficient as it looks up all pixels from $B$ to find a best match for a single pixel in $A$.
To make attention-based correspondence estimation feasible at higher-resolution, we propose an efficient sparse attention module. As shown in Fig.~\ref{fig:sparse_attention}, our attention module consists of feature map upsampling, key index sampling and sparse attentive warping to transfer visual content up to $512\times512$ resolution.
}

\noindent \textbf{Feature Map Upsampling.} \quad
\zht{To capture rich spatial details for transfer, we generate high-resolution cross-domain feature maps $u_x$ and $u_c$ from the low-resolution inputs $f_x$ and $f_c$. In particular, we apply spatially adaptive normalization~\cite{spade} to sequentially upsample the input feature until resolution $H\times W$:}
\begin{align}
\label{eq:feature_upsample}
\begin{aligned}
    u_x = U_x(f_x, c) \quad\text{and}\quad  u_c = U_c(f_c, x),
\end{aligned}
\end{align}
where $U_x(f_x;c)$ and $U_c(f_c, x)$ are two SPADE~\cite{spade} feature upsamplers that take $c$ or $x$ as guidance to upsample $f_x$ or $f_c$. We perform channel-wise normalization at the end of $U_x$ and $U_c$, such that the output features at every locations are normalized to have zero mean and unit $l_2$ norm.

\noindent \textbf{Sparse Attention.} \quad
Motivated by the sparsity of correspondences, 
for each query $q$, we constrain the search space of keys $p$ to be in a sparse key set $\mathbb{K}_q$ where $|\mathbb{K}_q|\ll H W$. Therefore, Eq.~\ref{eq:warping_dense} becomes: 
\begin{align}
\label{eq:warping_sparse}
\begin{aligned}
    \displaystyle
    r(q) &= \sum_{p \in \mathbb{K}_q} a_{q p} x(p),
\end{aligned}
\end{align}
\zht{where $p$ iterates over the sparse key set $\mathbb{K}_q$, ${a}_{q p} = {\gamma e^{s_{q p}}}/{\sum_{p \in \mathbb{K}_q} {\gamma e^{s_{q p}}}}$ is the normalized weight computed from $s_{q p}=\gamma \langle u_c(q),u_x(p) \rangle$, the high-resolution feature similarities. Next}, we propose a key index set sampling step to generate $\mathbb{K}_q$ and a sparse attentive warping module that efficiently aggregates pixels over $\mathbb{K}_q$.

\noindent \textbf{Key Index Sampling.} \quad
Motivated by a randomized correspondence algorithm, PatchMatch~\cite{patchmatch}, we propose a key index sampling step to exploit the spatial coherency of feature correspondences for efficient index set generation.
Specifically, for each query position $q$, we append a randomly initialized particle $\mathbf{t}_q=(t_q^x, t_q^y)$ to the index set $\mathbb{K}_q$ as initialization, namely $\mathbb{K}_q=\{\mathbf{t}_q\}$. Moreover, we use notation $\mathbb{P}_q$ to denote a ``search pool'' for each location $q$.
To ensure that $\mathbb{K}_q$ covers the best matches, we alternate between a \textit{initialize-propagate-evaluate} step that updates $\mathbf{t}_q$ with a better matched particle and an \textit{accumulation} step that appends the current particle $\mathbf{t}_q$ to $\mathbb{K}_q$:
\begin{enumerate}
    \item Initialize: to update $\mathbf{t}_q$ with better matching, a search pool $\mathbb{P}_q$ is initialized at each location $q$ with $\mathbf{t}_q$ and $k$ random particles sampled in the neighboring space of $\mathbf{t}_q$, resulting in  $\mathbb{P}_q=\{\mathbf{t}_q\}\cup \{{\mathbf{t}_q+w\mathbf{R}^{(i)}}\}$, where $\mathbf{R}^{(i)}$ is a random vector uniformly sampled in $[-1,1]\times[-1,1]$, $w$ is a predefined window size and $0\leq i<k$;
    \item Propagate: to leverage spatial coherency, search pool $\mathbb{P}_{q^\prime}$ from $q^\prime\in N(q)$, the adjacent pixel of $q$ are propagated to $\mathbb{P}_q$, resulting in: $\mathbb{P}_q \gets \mathbb{P}_q\cup \bigcup_{q^\prime\in N(q)} \mathbb{P}_{q^\prime}$;
    \item Evaluate: particles from $\mathbb{P}_q$ with the maximal feature matching score is used to update $\mathbf{t}_q$, resulting in $\mathbf{t}_q \gets \argmax_{\mathbf{t} \in\mathbb{P}_q} s_{q t}$
    \item Accumulate: the current match $\mathbf{t}_q$ is appended to $\mathbb{K}_q$: $\mathbb{K}_q \gets \mathbb{K}_q \cup \{\mathbf{t}_q\}$.
\end{enumerate}
The above procedure is repeated $N-1$ times where all the sub-steps are in parallel for each $q$. 
Fig.~\ref{fig:corr_visualize} shows that our procedure can sample diversified particles including random ones and most matched ones. 
To sample even more key indices, we initialize, update and append $M$ particles $\mathbf{t}_q^{(k)} (0<k\leq M)$ in parallel following the same procedure, resulting in a total of $M N$ index samples for each query.

It is shown that the above propagation step can be implemented via convolution~\cite{duggal2019deeppruner}. The other steps can be implemented by built-in tensor operations. 
Please refer to the appendix for the pseudo code of the sampling procedure.

\noindent \textbf{Sparse Attentive Warping.} \quad
The warping operation of Eq.~\ref{eq:warping_sparse} is defined on sparse and irregular grid $\mathbb{K}_q$. 
To enable efficient implementation, we propose the sparse attentive warping operation that transforms Eq.~\ref{eq:warping_sparse} to its equivalent form via deformable convolution~\cite{deformableConv}:
\begin{align}
\label{eq:deformable}
\begin{aligned}
    \displaystyle
    r_{x\rightarrow c}(q) &= \sum_{p \in \mathbb{K}_q} a_{q p} x(p) = \sum_{k=1}^{|\mathbb{K}_q|} x(q+\Delta{q_k}) A_{qk},
\end{aligned}
\end{align}
where $p$ enumerate all elements from $\mathbb{K}_q$, while $\Delta{q_k}=p_k-q$ and $A_{qk}=a_{q p_k}$ are the offset and modulation factor of the deformable convolution kernel defined in~\cite{deformableConv}, respectively. Note that the deformable kernel weights in Eq.~\ref{eq:deformable} are set to $1$ and thus being omitted. 

\noindent \textbf{Training Scheme.} \quad
\zht{
We follow the training scheme of~\cite{cocosnet} to train our sparse attention module. 
Specifically, we append an additional SPADE generator~\cite{spade} to the sparse attention module to refine the warped image.
However, as the SPADE generator is designed for semantic image synthesis rather than local editing, after the training, we discard the SPADE generator and keep the sparse attention module only for the next section.
During the training, the loss defined on generated image and other losses will be backpropagated to jointly update feature extractors of the sparse attention module and the generator. Although the key index sampling procedure is non-differentiable, the warped output $r$ in Eq.~\ref{eq:warping_sparse} is differentiable w.r.t. feature maps $f_x$ and $f_c$. Gradients can thus propagate through the sparse attention module for training.
}

\zht{
We denote the synthesized output that follows the style of $x$ and the layout of $c$ as $y_{x\rightarrow c}$.
Following~\cite{cocosnet}, a data pair $(x_0,c_0)$ is sampled from the ADE20k dataset. Next, a new pair $(x_1, c_1)$ is spatially augmented from $(x_0,c_0)$ and $y_{x_1\rightarrow c_0}$ is synthesized by the model. Moreover, an image with similar semantics to $x_0$ and its label map, denoted by $(x_2, c_2)$, is retrieved and $y_{x_2\rightarrow c_0}$ is synthesized.
Similar to~\cite{cocosnet}, our training objective consists of domain alignment loss, feature matching loss, perceptual loss~\cite{perceptual} and contextual loss~\cite{contextual}.
To stabilize training, we also impose regularization on the downsampled feature maps, as described in the Appendix.
}


\noindent \textbf{\zht{Customization for Local Editing.}} \quad
Our sparse attention module is trained and applicable to the global alignment task of~\cite{cocosnet}.
\zht{However, the global correspondence does not take the contextual constraint into account for the task of local layout manipulation, causing inconsistency at the boundaries of edited regions. Moreover, misalignment can be further reduced by encouraging correspondence smoothness and adopting the following customization to the key index sampling step:}
i) to avoid semantic mismatching in the inference step, we add a large penalization constant to $s_{q t}$ if $q$ and $t$ belong to different semantic regions, ii) in the sampling step, we exponentially decay the window size $w$ as the iteration number increases, and iii) we repeat more propagate-evaluation steps in each iteration to enforce correspondence smoothness. The optimized procedure helps to improve the warping quality for local editing.

\vspace{-2mm}
\subsection{Semantic Layout Manipulation Network}
\label{sec:network}
As shown in Fig.~\ref{fig:pipeline}, our semantic layout manipulation network takes an image $x_0$ and its layout $c_0$, an edited layout $c_1$ and a mask $m$ which indicates the region to-be-edited as inputs. Our generator denoted by $G$ aims to hallucinate new pixels inside the mask to generate the edited image $y$:
\begin{align}
\label{eq:generator}
\begin{aligned}
    y = G(x_0,c_0,c_1,m) \odot m + x_0 \odot (1-m),
\end{aligned}
\end{align}
whereas the generated contents inside the mask should be i) aligned to the layout $c_1$ and ii) coherent with the neighboring content $x_0 \odot (1-m)$. 

\noindent \textbf{High-resolution Local Alignment.} \quad
Aiming at transferring visual details from the input image to handle drastic layout change, we first apply spatial alignment described in Sec.~\ref{sec:HR_attention} to generate an layout-aligned image $r_{x_0\rightarrow c_1}$ that is consistent to $x_0$ and follow the layout of $c_1$. The pixels outside the mask are then replaced by the real image, resulting in a \textit{locally warped} image:
\begin{align}
\label{eq:local_align}
\begin{aligned}
    x_{\text{warp}} = r_{x_0\rightarrow c_1} \odot m + x_0 \odot (1-m).
\end{aligned}
\end{align}
As the locally warped images $x_{\text{warp}}$ may contain appearance discontinuity due to misalignment, we next propose a generator that consists of a semantic encoder and coarse-to-fine decoders to refine the warped result.

\noindent \textbf{Semantic Encoder.} \quad
To better capture semantic and structure clue from the noisy output $x_{warp}$, we use a pretrained visual-semantic embedding (VSE) model~\cite{vse++} to extract multi-scale visual features $f^{(1)}, f^{(2)}, f^{(3)}, f^{(4)}  = \VSE(x_\text{warp})$.
Likewise, we extract layout embedding $f_{c_1}$ from the one-hot layout label map of $c_1$ using a $1\times 1$ convolution layer.

\zht{
\noindent \textbf{Coarse-to-fine Synthesis} \quad
As the warping result $x_\text{warp}$ may contain misalignment or halo artifact as shown in Fig.~\ref{fig:pipeline}, we propose a coarse-to-fine refinement network to refine $x_\text{warp}$. 
In particular, our coarse-stage network consists of a two-branch encoder that utilizes dilated convolution~\cite{dilatedconv} and contextual attention~\cite{deepfillv1} to propagate visual information from the neighboring regions of $x_0$. The two-branch encoder takes the concatenation $[x_\text{warp},f_{c_1},m,x_0\odot (1-m)]$ as input to extract multi-scale features. Furthermore, at each scale $i$ of the two-branch encoder, visual features at the same scale $f^{(i)}$ and the resized layout embedding $f_{c_1}$ are also concatenated to help propagating contextual information. After concatenating the two feature maps generated by the encoders, a coarse stage decoder that consists of multiple deconvolution layers sequentially upsample feature resolution and produce an coarse image prediction $y_{\text{coarse}}$.


Motivated by the coarse-to-fine pipeline~\cite{deepfillv1,xiong2019foreground} of inpainting models, we employ a refinement network based on SPADE~\cite{spade} to refine $y_{\text{coarse}}$. However, we feed the intermediate features from the coarse-stage decoder $f_{\text{if}}$ and $[x_\text{warp},y_{\text{coarse}},f_{c_1},m]$ as additional inputs to our refinement network, such that the contextual information from the coarse-stage network and inputs can be exploited.
}

\zht{
\noindent \textbf{Self-supervised Training} \quad
It is difficult to collect ground-truth triplets $(x_0,c_1,x_1)$ such that the edited groundtruth $x_1$ is aligned to the layout of $c_1$ while being coherent with neighboring content of $x_0$. However, assuming such $x_1$ exists, $x_0$ will only be involved in Eq.~\ref{eq:local_align} in our network to generate the warped image $r_{x_0\rightarrow c_1}$.\footnote{Whereas the neighboring content of the input image $x_0 \odot (1-m)$ can be the neighboring content of  $x_1$, namely $x_1 \odot (1-m)$.} This motivate us to  design a novel unsupervised scheme that relies on approximated warped content $r_{x_0\rightarrow c_1}$ instead of $x_0$ to generate the edited output.

Specifically, during training, we sample image-layout pairs $(x_1, c_1)$ from datasets and free-form mask $m$ using the procedure described in~\cite{deepfillv2}. Instead of collecting $x_0$, we apply random translation and flipping on $x_1$, resulting in augmented image $\widetilde{x}_1$. Since $x_0,x_1$ and $\widetilde{x}_1$ are coherent in content, we approximate $r_{x_0\rightarrow c_1}$ with $r_{\widetilde{x}_1\rightarrow c_1}$ whereas the locally warped image in Eq.~\ref{eq:local_align} is replaced by:
\begin{align}
\label{eq:local_align2}
\begin{aligned}
    x_{\text{warp}} = r_{\widetilde{x}_1\rightarrow c_1} \odot m + x_1 \odot (1-m).
\end{aligned}
\end{align}
We found that such a self-supervised scheme works well in general on arbitrarily edited layout. We argue that the generalization capacity of our model come from the drastic layout change introduced by flipping during training.

We first train the first-stage network using a combination of pixel-wise $\ell_1$ loss and perceptual loss~\cite{perceptual} to give coarse prediction. 
Subsequently, the second-stage network is jointly trained with the initially trained first-stage network using a combination of two-stage pixel-wise $\ell_1$ loss, perceptual loss~\cite{perceptual} $\mathcal{L}_{\text{prec}}$ and adversarial loss to give the final output:
\begin{align}
\label{eq:generator_loss}
\begin{aligned}
    \mathcal{L}_G = &\lambda_{\ell 1} \mathcal{L}_{\ell \text{1-stage1}} + 
                    \lambda_{L1} \mathcal{L}_{\ell \text{1-stage2}}+ \\ &\lambda_{\text{prec}} \mathcal{L}_{\text{prec}} - \lambda_{\text{adv}} \mathop{\mathbb{E}}_{z\sim p(z)}[D(y,c,m)],
\end{aligned}
\end{align}
where $\lambda$ represents the balancing terms. For discriminator, we apply the Markovian discrinimator~\cite{deepfillv2} with hinge loss:
\begin{align}
\label{eq:discriminator_loss}
\begin{aligned}
    \mathcal{L}_D = & \mathop{\mathbb{E}}_{z\sim p_{\text{data}}}[\sigma(1-D(x,c,m))] + \mathop{\mathbb{E}}_{z\sim p(z)}[\sigma(1+D(y,c,m))].
\end{aligned}
\end{align}
}

\vspace{-0.3in}
\section{Experiments}
\label{sec:experiment}
\subsection{Implementation Details}
\label{sec:experiment_implementation}
We train our high-resolution sparse attention model on the ADE20k dataset~\cite{ade20k} for 60 epochs.
We then train semantic layout manipulation model on the Places365~\cite{places2} dataset.
We generate semantic label maps on Places365 using the HRNets~\cite{hrnet} model pretrained on ADE20k dataset. During testing, we apply a refinement network~\cite{cascadepsp} to post-process the predicted labels maps.
To perform the key index sampling step, we set the max iteration $N=15$ and particle number $M=2$. We found that the total iteration number $N$ is important to the result. In particular, if $N$ is too small (e.g. $N=5$), the generated correspondence tend to be noisy and inconsistent. Moreover, much larger iterations does not bring significant benefit on performance. As the memory footprint of the sparse attention step linearly grows with respect to $MN$, we found $N=15$ and $M=2$ avoids using excessive GPU memory.
%

\vspace{-2mm}
\subsection{Comparisons with Existing Methods}
\label{sec:experiment_compare}
We compare our method with the state-of-the-art inpainting methods including  Deepfill-v2~\cite{deepfillv2}, Profill~\cite{profill}, Edge-connect~\cite{edgeconnect}, MEDFE~\cite{liu2020rethinking}, and PatchMatch~\cite{patchmatch}, semantic layout manipulation methods including SESAME~\cite{sesame}, and global layout editing method CoCosNet~\cite{cocosnet} on the ADE20k and Places365 validation sets. For those methods, we feed label maps (or edge map of the label maps) whenever possible. 

\begin{figure}[t]
	\centering
	\includegraphics[width=1.0\linewidth]{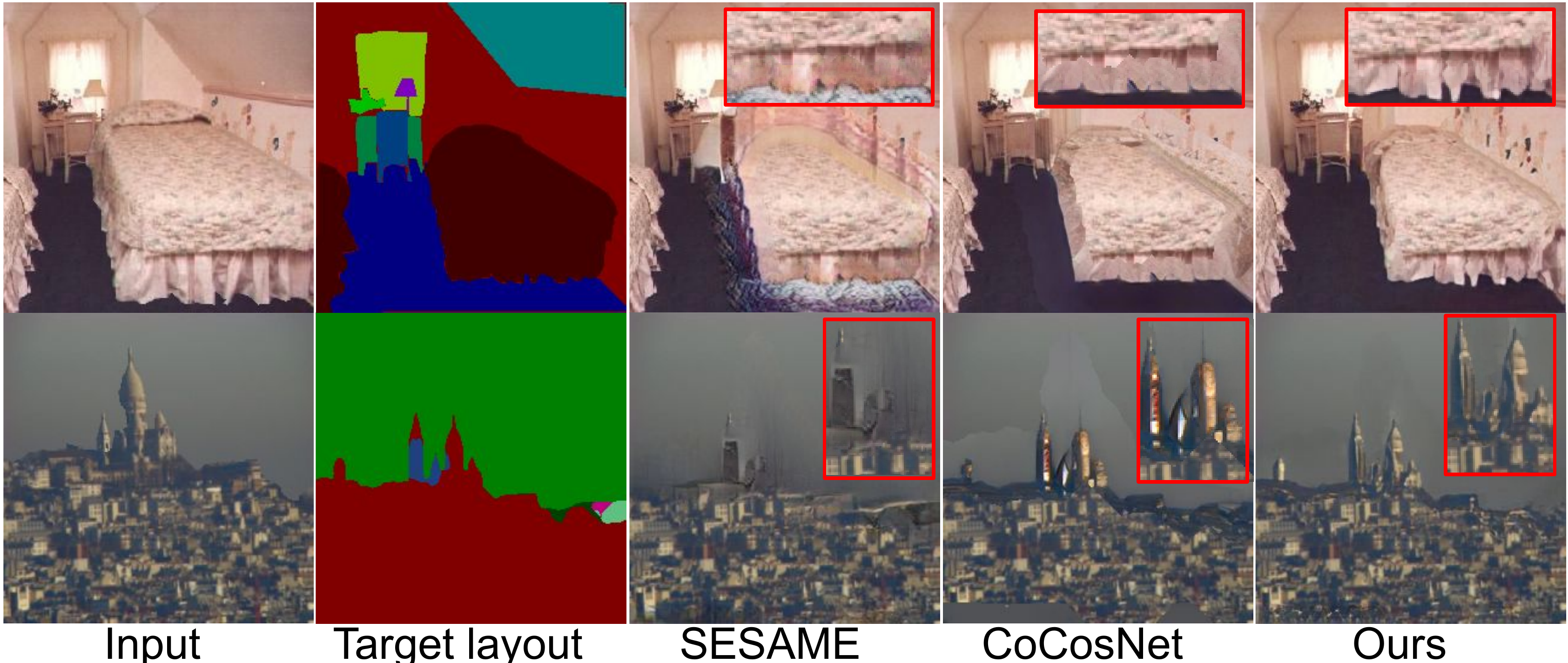}
	\caption{
		\textbf{Qualitative comparisons on the manipulation task.} We show from left to right the input image, target layout, and the results of SESAME~\cite{sesame}, CoCosNet~\cite{cocosnet} and our model, respectively. Best viewed by zoom-in on screen.
	}
	\label{fig:main_compare}
\end{figure}

\zht{
\noindent \textbf{Quantitative Evaluation} \quad
We use \textit{reconstruction} and \textit{manipulation} tasks to evaluate our method. The reconstruction task is adapted from  guided inpainting~\cite{deepfillv2,sesame} to reconstruct groundtruth images from masked images and guided inputs (such as label maps or edge maps).
The manipulation task aims to generate a manipulated image using a locally edited label map. To generate such an evaluation set, we randomly sample a segment from the evaluation images and apply random affine transformation on the segment to generate the image before editing. Fig.~\ref{fig:main_compare} shows the resulting synthetic data. For fair comparisons on the reconstruction task, we modify our warping stage 
such that the attention weights for the masked region are set to 0 and pixels inside the mask are ignored in the attention-based alignment stage.
}

\begin{table}[]
	\caption{
		\textbf{Quantitative evaluation on the reconstruction and manipulation tasks.} Higher scores are better for the metrics with uparrow ($\uparrow$), and vice versa. 
	}
	\centering
	\resizebox{\columnwidth}{!}{
	\begin{tabular}{ |l|c|c|c|c|c|c|}
		\hline
		\hline
		Methods                             &$\ell_1$ err. $\downarrow$ &	PSNR$\uparrow$ & SSIM$\uparrow$ &  LPIPS$\downarrow$	& FID$\downarrow$ & $\mathcal{L}_{\text{style}}$~\cite{gatys2015neural}$\downarrow$\\
		\hline
		\multicolumn{7}{|l|}{\em ADE20k reconstruction}\\
		\hline
		Deepfill\_v2~\cite{deepfillv2}&  0.02959&  21.762&  0.844 & 0.305& 85.73 &4.440e-06\\
		MEDFE~\cite{liu2020rethinking}&  0.03825&  20.730&  0.808& 0.413& 167.12 &1.528e-05\\
		Profill~\cite{profill}&  0.02875&  22.405&  0.848& 0.317& 97.78&8.183e-06\\
		Edge-connect~\cite{edgeconnect}&  0.04512&  20.179&  0.742 & 0.284& 80.96&6.061e-06\\
		SESAME~\cite{sesame}&  0.06032&  18.505&  0.720 & 0.319& 66.45&1.300e-05\\
		Ours&  \best{0.02486}&  \best{23.173}&  \best{0.863} & \best{0.217}& \best{49.42}&\best{2.800e-06}\\
		\hline
		\multicolumn{7}{|l|}{\em ADE20k manipulation}\\
		\hline
		SESAME~\cite{sesame}&  0.03561&  20.825&  0.834 & 0.178& 47.40& 5.281e-06\\
		CoCosNet~\cite{cocosnet}&  0.03065&  24.063&  0.837 & 0.165& 98.13&\best{1.397e-06}\\
		Ours&  \best{0.02827}&  \best{22.672}&  \best{0.859} & \best{0.136}& \best{28.07}&2.597e-06\\
		\hline
		\multicolumn{7}{|l|}{\em Places365 reconstruction}\\
		\hline
		Deepfill\_v2~\cite{deepfillv2}&  0.06012&  18.145&  0.688 & 0.297& 104.00 &\best{4.082e-06}\\
		MEDFE~\cite{liu2020rethinking}& 0.07053&  17.498&  0.652& 0.402& 160.86 &3.672e-05\\
		Profill~\cite{profill}& 0.05597&  18.758&  0.699 & 0.298& 103.17&1.211e-05\\
		Edge-connect~\cite{edgeconnect}&  0.05448&  19.158&  0.694 & 0.286& 97.21&6.517e-06\\
		SESAME~\cite{sesame}&  0.06050&  18.586&  0.684 & 0.326& 119.72&1.406e-05\\
		CoCosNet~\cite{cocosnet}&  0.07759&  16.841&  0.644 & 0.356& 134.80&1.235e-05\\
		Ours&   \best{0.05209}&  \best{19.478}&  \best{0.714} & \best{0.254}& \best{84.68} &5.811e-06\\
		\hline
		\multicolumn{7}{|l|}{\em Places365 manipulation}\\
		\hline
		SESAME~\cite{sesame}&  0.02617&  22.957&  0.879 & 0.118& 49.24&1.831e-06\\
		CoCosNet~\cite{cocosnet}&  0.03050&  22.979&  0.873 & 0.126& 47.47&1.397e-06\\
		Ours &  \best{0.02500}&  \best{24.252}&  \best{0.888}& \best{0.102} & \best{31.35}& \best{6.801e-07}\\
		\hline
		\hline
	\end{tabular}
	}
	\label{tab:main_experiment_reconstruct}
\end{table}

For quantitative measurement, we apply average L1 Error ($\ell_1$), PSNR and SSIM~\cite{ssim} as the low-level metrics and Frchet Inception Distance (FID)~\cite{fid}, Perceptual Image Patch Similarity Distance (LPIPS)~\cite{lpips} and the style similarity loss $\mathcal{L}_{\text{style}}$ of~\cite{gatys2015neural} as perceptual-level metrics. Results on manipulation and reconstruction tasks are 
in Table~\ref{tab:main_experiment_reconstruct}.
For both tasks, our method performs the best according to most metrics. Additionally, we observe that the semantics-guided approaches (SESAME~\cite{sesame}, CoCosNet~\cite{cocosnet} and ours) in general outperforms other inpainting approaches.

\noindent \textbf{Qualitative Evaluation} \quad
Fig.~\ref{fig:main_compare} presents visual comparisons on the image manipulation task where input images are manipulated in regions indicated by the red rectangles. 
SESAME~\cite{sesame} generates coherent colors but produces structural artifacts (e.g. the tower in the last row), as the masked pixel is not leveraged by warping.
CoCosNet~\cite{cocosnet} better transfers appearance via alignment. However, CoCosNet hallucinates wrong objects (tower in the last row) as warping is performed in lower resolution and may contain error.
Moreover, CoCosNet often generates pixels that have incoherent colors with their surrounding due to the lack of local refinement step. In contrast, our method better preserves texture details of input images, generating more coherent results.

\begin{figure}[]
	\centering
	\includegraphics[width=1\linewidth]{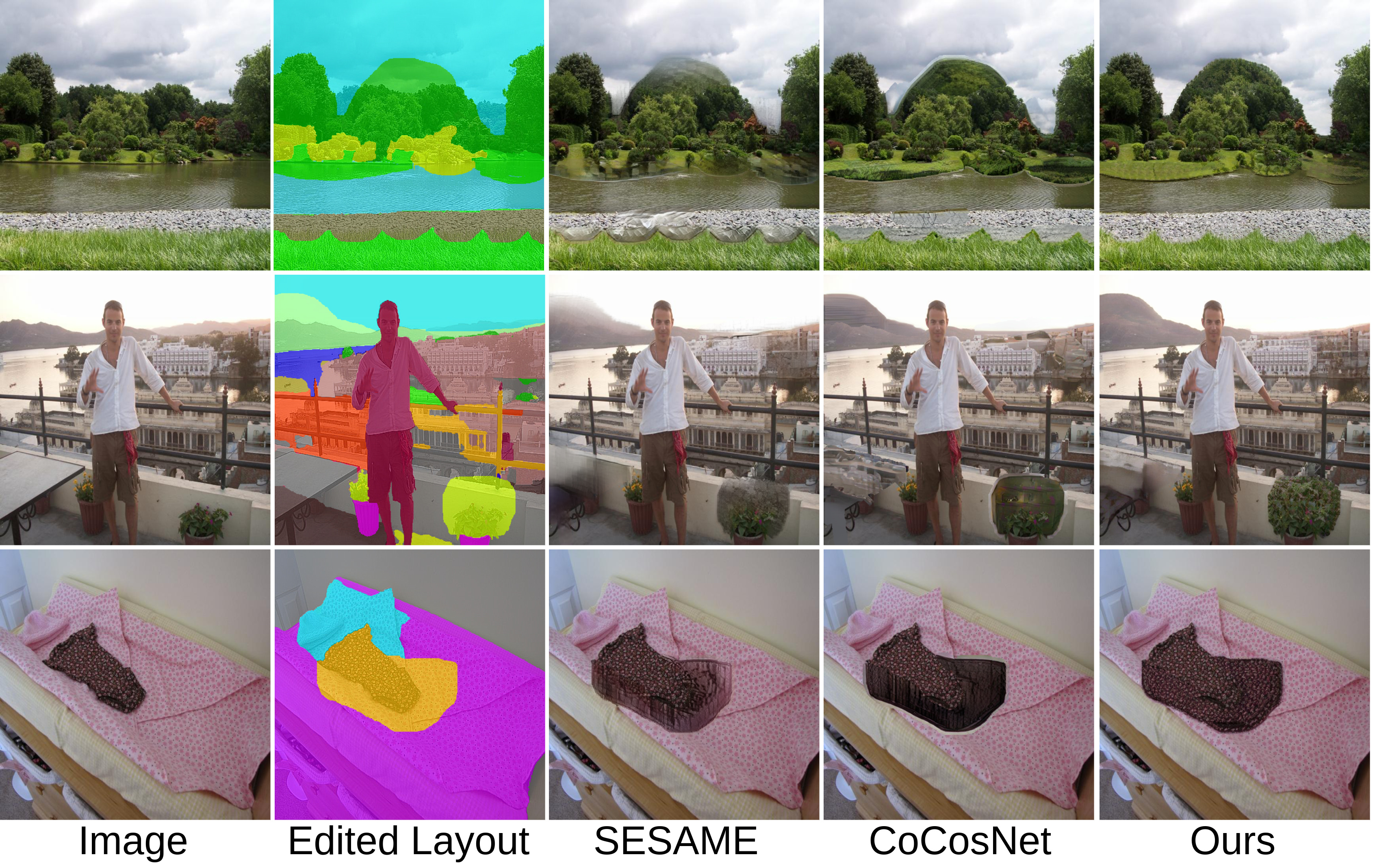}
	\caption{
		\textbf{Qualitative comparisons on the real manipulation task.} Best viewed by zoom-in on screen.
	}
	\label{fig:main_compare_handcraft}
\end{figure}

\begin{figure*}[t]
	\centering
	\includegraphics[width=1\linewidth]{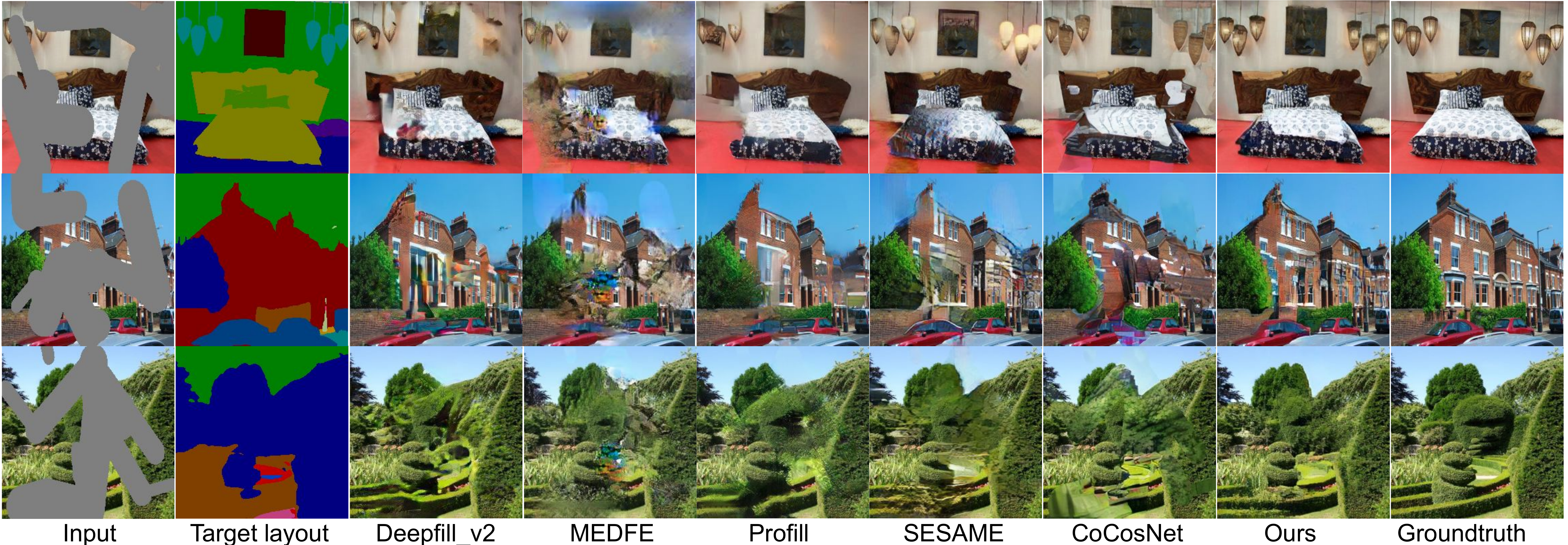}
	\caption{
		\textbf{Qualitative comparisons on the reconstruction task.} We show from left to right the input image, target layout, and the results of Deepfill\_v2~\cite{deepfillv2}, MEDFE~\cite{liu2020rethinking}, Profill~\cite{profill}, SESAME~\cite{sesame}, CoCosNet~\cite{cocosnet} and our model, respectively. Best viewed by zoom-in on screen.
	}
	\label{fig:main_compare_inpaint}
\end{figure*}

Fig.~\ref{fig:main_compare_inpaint} shows the comparisons from the reconstruction task.
Benefiting from the semantic layout guidance, SESAME~\cite{sesame}, CoCosNet~\cite{cocosnet} and our approach better reconstruct the image layout. Among them, our approach better preserves details from the inputs (e.g. texture of bed sheet, building). In addition, Fig.~\ref{fig:main_compare_handcraft} compares the image manipulation results on real data, where our method achieves the best visual results.

\begin{figure}[t]
	\centering
	\includegraphics[width=1.0\linewidth]{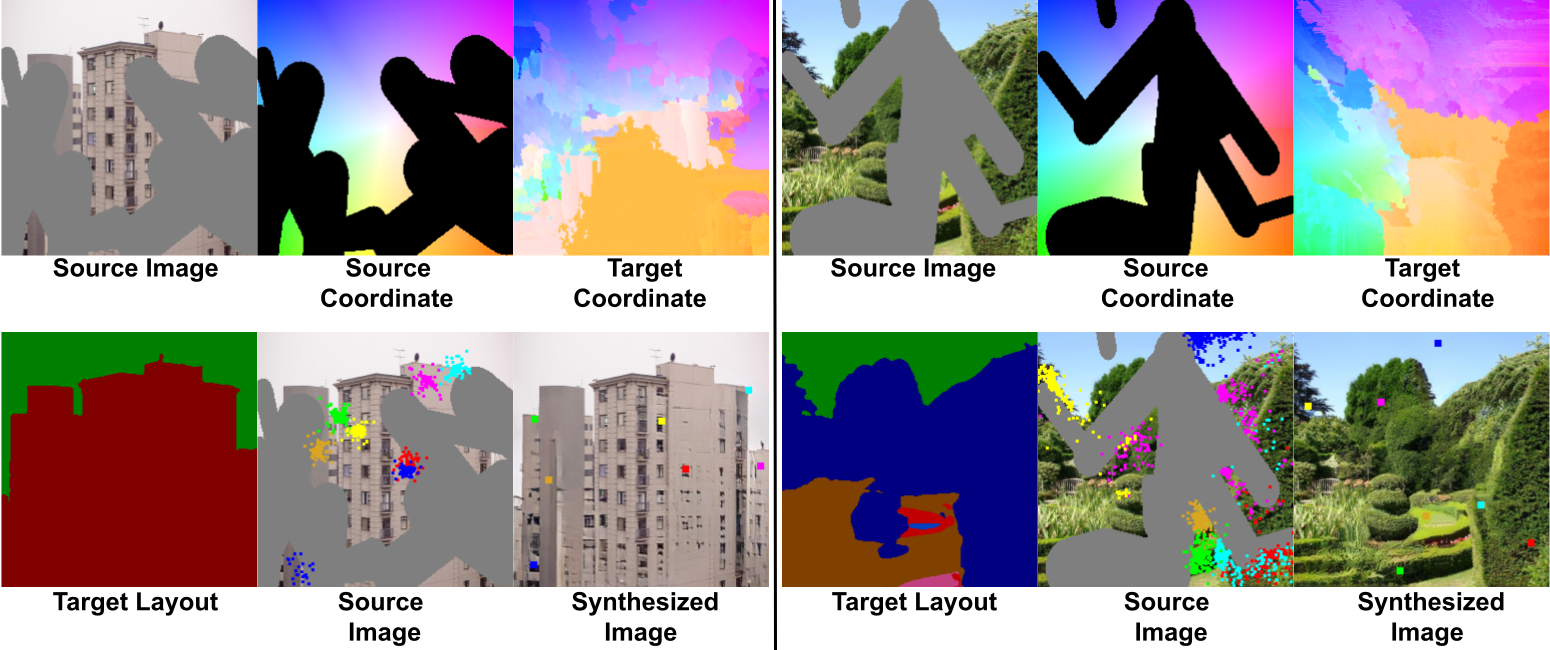}
	\caption{
		\textbf{Visualization of the correspondence generated by our model.} Best viewed (e.g. local textures) with zoom-in on screen.
	}
	\label{fig:correspondence}
\end{figure}

\begin{figure}[]
	\centering
	\includegraphics[width=1.0\linewidth]{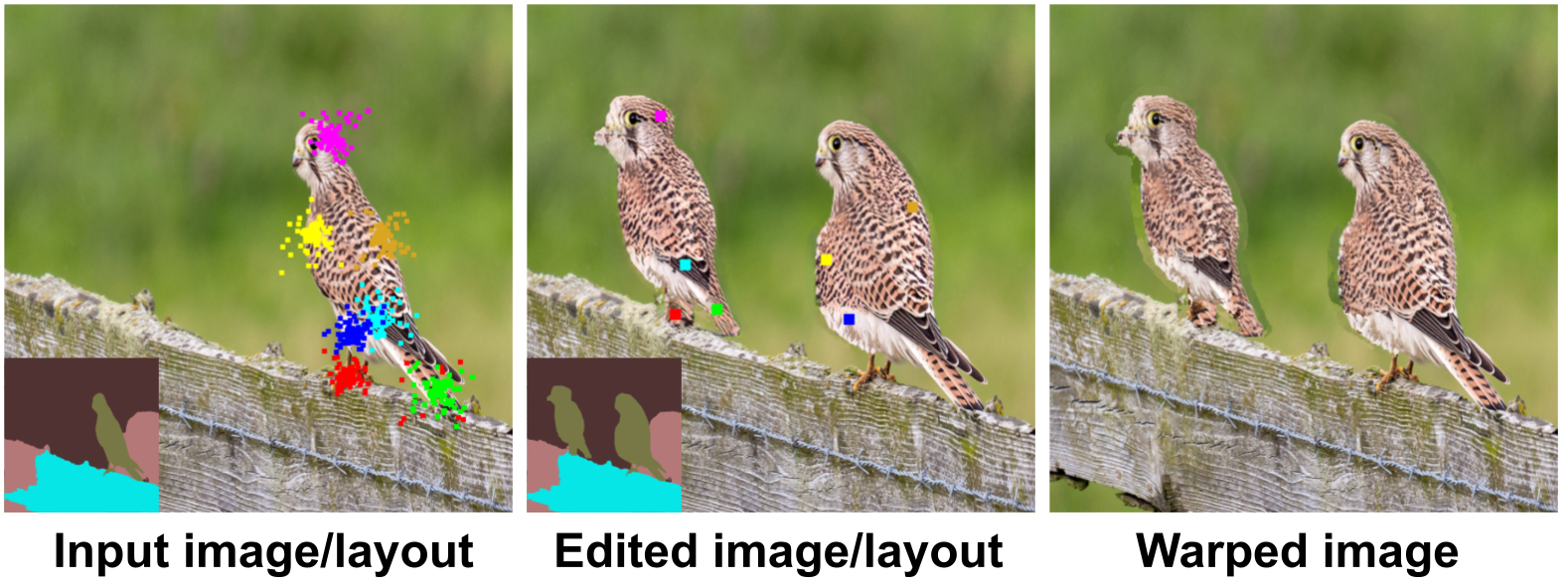}
	\caption{
		\textbf{Correspondence generated by our sparse attention module.} The colored points in the left and middle figures represent the query position and the sampled sparse keys, respectively. The left figure visualizes the warped image by the sparse attention module.
	}
	\label{fig:corr_visualize}
\end{figure}

Fig.~\ref{fig:corr_visualize} shows an example visualization of the correspondences generated by the proposed sparse attention module.
Using the proposed selective sampling procedure, our sparse attention module can effectively locate regions of interest to enable efficient attention computation. In addition, the synthesis module can correct the inconsistent textures and halo artifacts of the warped image (Fig.~\ref{fig:corr_visualize} left) and produce the refined output (Fig.~\ref{fig:corr_visualize} middle).
In addition, as shown in Fig.~\ref{fig:correspondence}, our correspondence can borrow correct semantic such as windows, sky, and tree from the source image to the target layout as shown in the second row and when there are multiple plausible correspondence, it transfers appearance from a corner of building/tree of the source image to the corner of building/tree of the target layout.

\begin{figure}[t]
	\centering
	\includegraphics[width=1.0\linewidth]{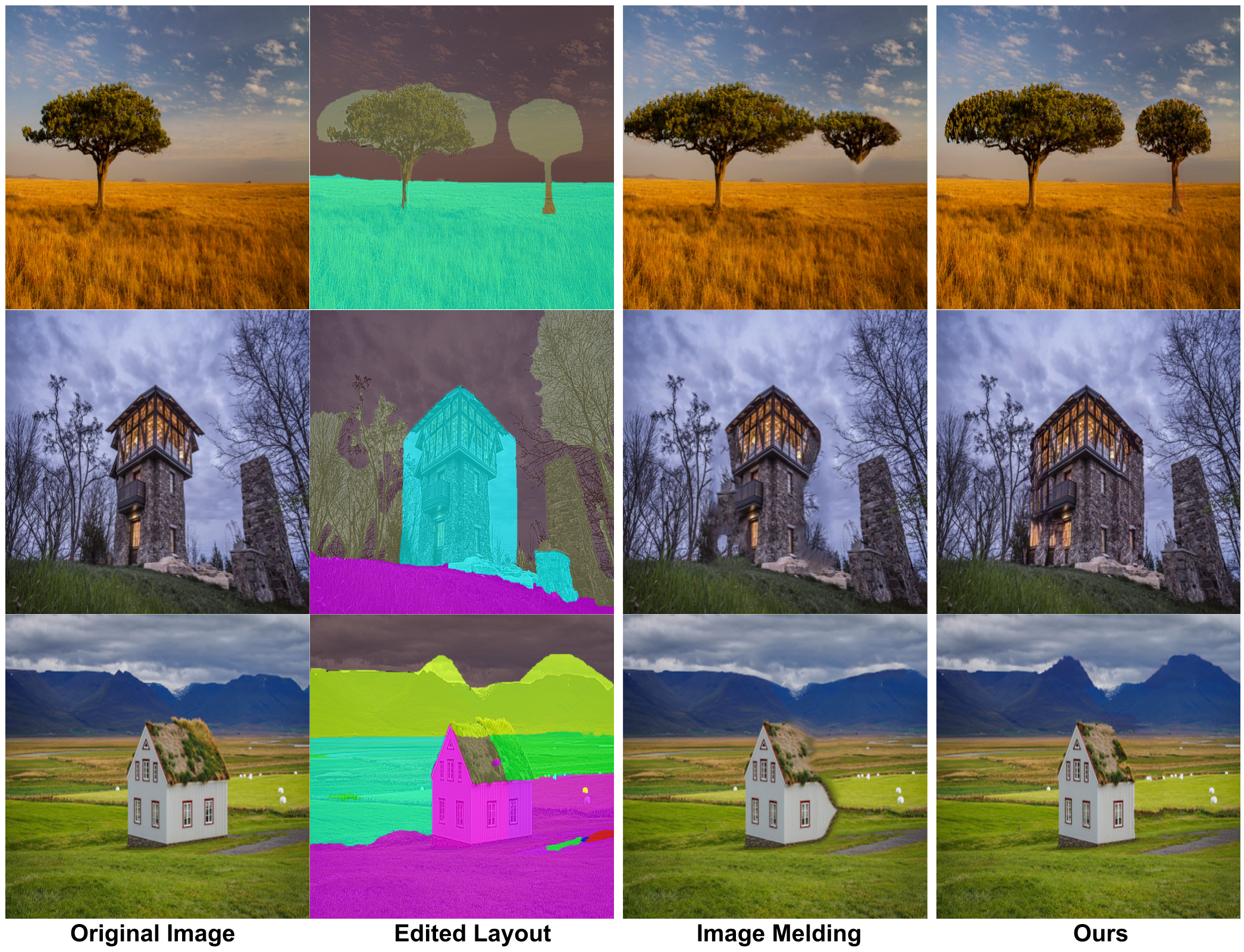}
	\caption{
		\textbf{Qualitative comparisons to image melding~\cite{image_melding}.}
	}
	\label{fig:image_melding}
\end{figure}

\begin{figure}[t]
	\centering
	\includegraphics[width=1.0\linewidth]{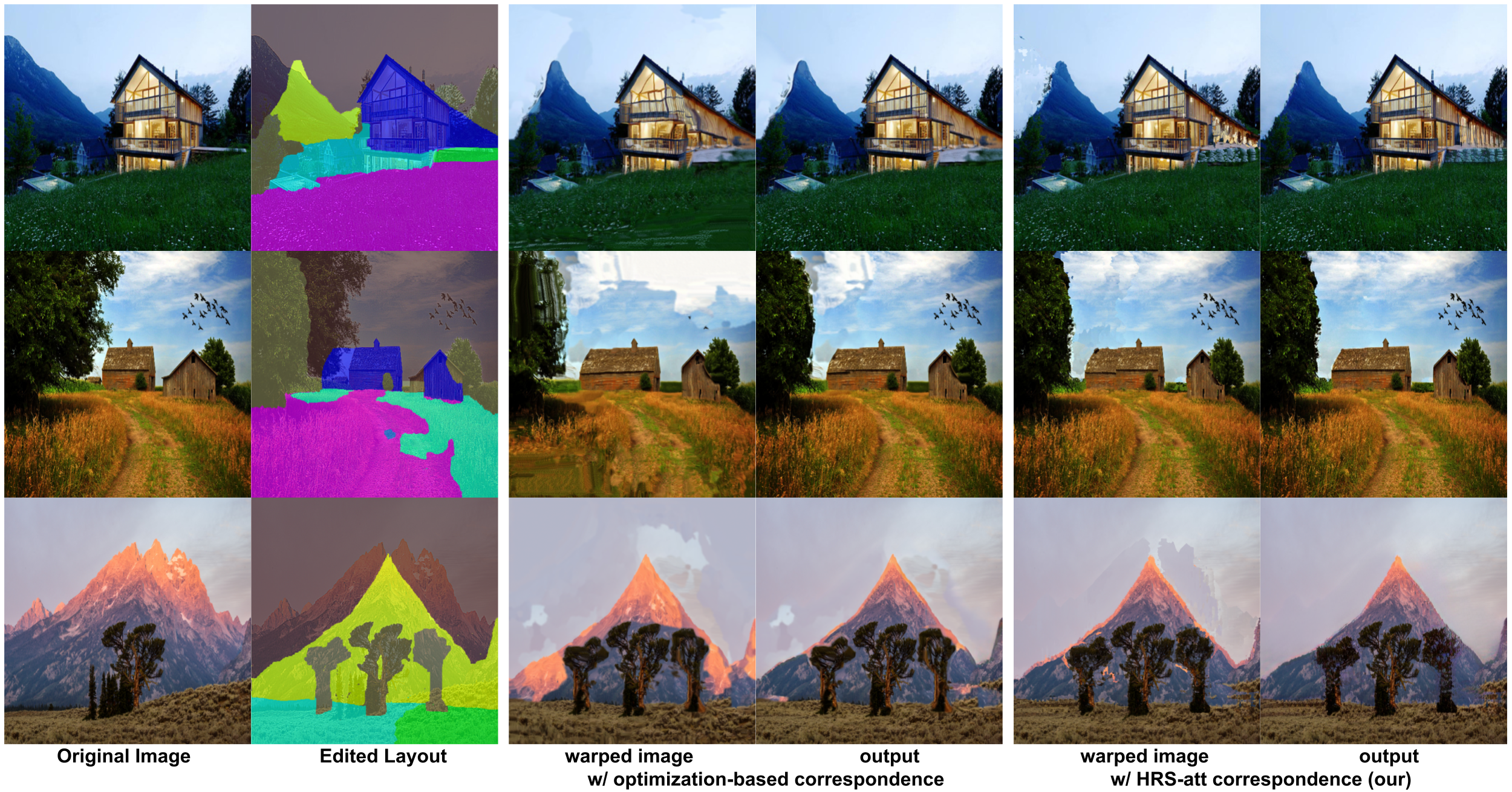}
	\caption{
		\textbf{Qualitative comparisons to deep image analogy~\cite{deep_image_analogy}.} Column 2 and 3 are the warped images and final outputs of~\cite{deep_image_analogy}. Column 4 and 5 are our warped images and final outputs. Best viewed (e.g. local textures) with zoom-in on screen.
	}
	\label{fig:optimization}
\end{figure}

In addition, we introduce the qualitative comparison to other optimization-based approaches including deep image analogy~\cite{deep_image_analogy} and image melding~\cite{image_melding}. Specifically, as deep image analogy does not support cross domain correspondence, we apply our trained model to extract cross domain features at resolution $512\times512$ and resize the feature to resolution $256, 128, 64, 32, 16$ to produce the multi-scale feature pyramid. Then, we adopt the bidirectional patchmatch and feature deconvolution~\cite{deep_image_analogy} to optimize the dense correspondence and the warped image. Finally, we leverage our synthesis network to generate the final output from the warped image. As shown in Fig.~\ref{fig:optimization}, the warped images generated via optimization (3'rd column) are distorted while the final outputs (4'th column) are inconsistent in comparison to our warped images (5'th column) and final outputs (6'th row).
Moreover, deep image analogy~\cite{deep_image_analogy} takes 10.5 minites to compute correspondence while our approaches takes 0.4 seconds to compute correspondence.
In addition, Fig.~\ref{fig:image_melding} presents the visual comparison to image melding~\cite{image_melding}. While image melding produces satisfactory local textures, our approach generate better global semantics structure with realistic local details thanks to the cross-domain and high-resolution warping scheme.

\subsection{User Study}
To better evaluate visual quality of our method, we conduct a user study and show the result in Table~\ref{tab:sup_ablation}.  Specifically, we randomly select 15 images from the Place365 validation set for manipulation and reconstruction, respectively. We show results of each image to 7 users and ask them to select the best results. Finally, we collect 105 votes from all users for each dataset. Results in the table show that our method receives the majority of the votes on both manipulation and reconstruction tasks.
%

\begin{table}[]
	\caption{User preference for the results of all methods. 
	}
	\centering
	 	\resizebox{\columnwidth}{!}{
	\begin{tabular}{ |l|c|l|c|}
		\hline
		\hline
		\multicolumn{2}{|c}{\em Manipulation} & \multicolumn{2}{c|}{\em Reconstruction}\\
		\hline
		Methods&Preference Votes&Methods&Preference Votes\\
		\hline
		SESAME~\cite{sesame}&  0 & Deepfill\_v2~\cite{deepfillv2}& 9\\
		CoCosNet~\cite{cocosnet}&  7& Edge-connect~\cite{edgeconnect}& 1\\
		Ours&  \best{98} & MEDFE~\cite{liu2020rethinking}& 0\\
&&Profill~\cite{profill}& 13\\
&&SESAME~\cite{sesame}&  2\\
&&CoCosNet~\cite{cocosnet}&  5\\
&&Ours&   \best{75}\\
		\hline
		\hline
	\end{tabular}
	 	}
	\label{tab:sup_ablation}
\end{table}

\vspace{-0.1in}
\subsection{Ablation Study}
We perform a set of ablation experiments to show the importance of the network design and usage of the high-resolution correspondence. Results of the ablations are shown in Table~\ref{tab:ablation}. To study the network design, we train four variants of our model: i) \textit{our w/o SE} that removes the \textbf{s}emantic \textbf{e}ncoder branch from the generator, ii) \textit{our w/o CA} that removes the \textbf{c}ontextual \textbf{a}ttention branch from the two-stage encoder, iii) \textit{our w/o dilate} that removes the dilated convolution branch branch from the two-stage encoder, and iv) \textit{our w/o IF} that does not feed the \textbf{i}ntermediate \textbf{f}eatures from coarse-stage decoder to the refinement network. From the results, our full model achieves the best performance on the metrics that reflect structure-level similarity (SSIM) and semantic similarity (LPIPS, FID). We also observe that removing the contextual attention or dilated convolution branch impact performances the most suggesting that contextual propagation is important for local editing.

To study the effect of the high-resolution correspondence, we adopt two variants of locally warped image $x_{\text{warp}}$ from Eq.~\ref{eq:local_align} to train models: i) \textit{our w/o corres} set $r_{x_0\rightarrow c_1}$ to $0$, resulting in a model that does not relies on correspondence, and ii) \textit{our w/ LR corres} set $r_{x_0\rightarrow c_1}$ to the low-resolution warped image of CoCosNet~\cite{cocosnet}. From the results, we observe that the quality of correspondence impact the final performance. Benefiting from the high-resolution correspondence, our full model achieves the best performance while \textit{our w/ LR corres} achieves better performance than \textit{our w/o corres} due to the usage of correspondence. Notably, sharing similar input as SESAME~\cite{sesame}, \textit{our w/o corres} substantially outperforms SESAME~\cite{sesame}. More visual comparisons are provided in the Appendix.


\begin{table}[]
	\caption{Quantitative evaluation of the ablation models on the Places365 manipulation task.
	}
	\centering
	\resizebox{\columnwidth}{!}{
	\begin{tabular}{ |l|c|c|c|c|c|c|}
		\hline
		\hline
		Methods                             &$\ell_1$ err. $\downarrow$ &	PSNR$\uparrow$ & SSIM$\uparrow$ &  LPIPS$\downarrow$	& FID$\downarrow$ & $\mathcal{L}_{\text{style}}$~\cite{gatys2015neural}$\downarrow$\\
		\hline
		ours full &  \best{0.02500}&  \best{24.252}&  \best{0.888}& \best{0.102} & \best{31.35}& 6.801e-07\\
		\hline
		ours w/o SE&  0.02525&  23.940&  0.887& 0.104& 33.51& \best{4.875e-07}\\
		ours w/o CA&  0.02629&  23.879&  0.885&  0.104& 32.18& 7.896e-07\\
		ours w/o dilate&  0.02602&  24.036&  0.885&  0.105& 32.43& 7.760e-07\\
		ours w/o IF& 0.02534&  24.100&  0.887& 0.105 & 33.02& 6.731e-07\\
		\hline
		ours w/o corres&  0.02764&  22.999&  0.877&  0.112& 38.61& 1.032e-06\\
		ours w/ LR corres& 0.02619&  23.751&  0.885& 0.104 & 32.65& 8.252e-07\\
		\hline
		\hline
	\end{tabular}
	}
	\label{tab:ablation}
\end{table}

\begin{figure}[h]
	\centering
	\includegraphics[width=.9\linewidth]{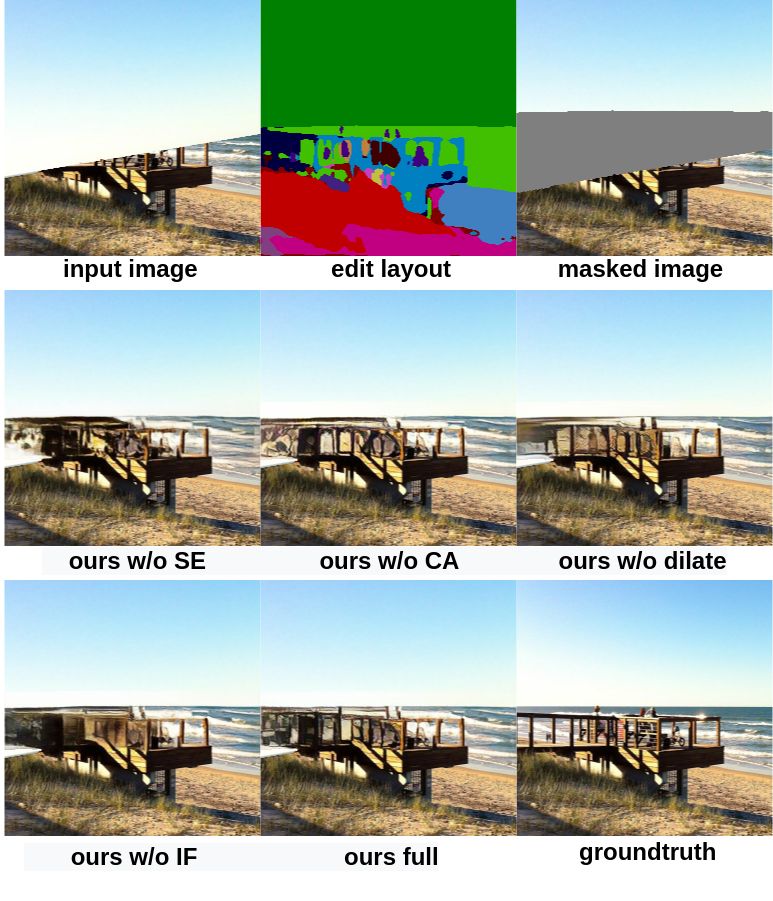}
	\caption{\textbf{Qualitative comparisons of ablation models (on the Places365 dataset).} Best viewed (e.g. local textures) with zoom-in on screen.}
	\label{fig:supp_compare_ablation1}
\end{figure}
\begin{figure}[h]
	\centering
	\includegraphics[width=.9\linewidth]{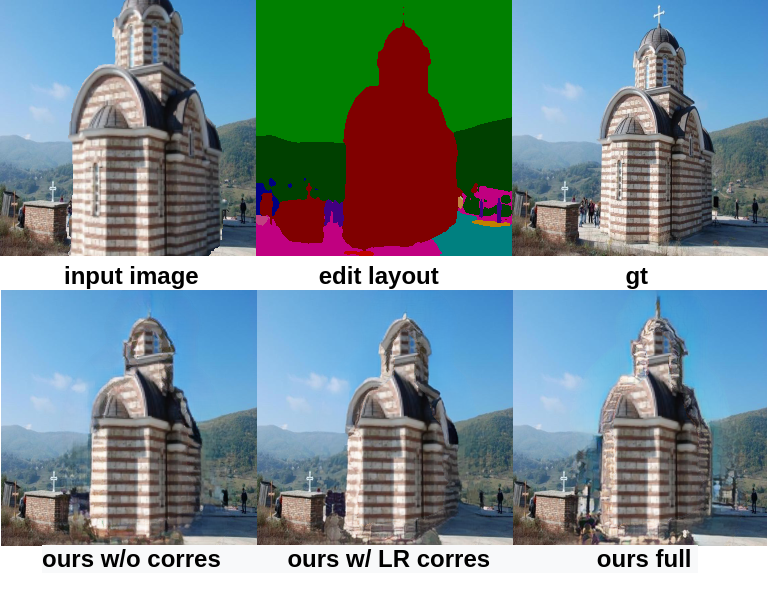}
	\caption{\textbf{Qualitative comparisons of ablation models (on the Places365 dataset).} Best viewed (e.g. local textures) with zoom-in on screen.}
	\label{fig:supp_compare_ablation2}
\end{figure}

We also present the visual comparisons of the ablation models in Fig.~\ref{fig:supp_compare_ablation1} and Fig.~\ref{fig:supp_compare_ablation2}. From Fig.~\ref{fig:supp_compare_ablation1}, our full model synthesizes realistic objects. In contrast, \textit{ours w/o SE} generates artifacts around the object. In addition, without the intermediate features, \textit{ours w/o IF} generates blurry textures and it sometimes discards structure details. From Fig.~\ref{fig:supp_compare_ablation2}, we observe that the quality of correspondence impact the model performance. Specifically, \textit{ours w/o corres} sometimes cannot generate realistic structure. Benifiting from correspondence, \textit{ours w/ LR corres} and our full model achieves better results. However \textit{ours w/ LR corres} often generate blurry local textures while our model can generate sharp texture.

\subsection{Discussion on Generalization to Layout Change}
\begin{figure}[t]
	\centering
	\includegraphics[width=1.0\linewidth]{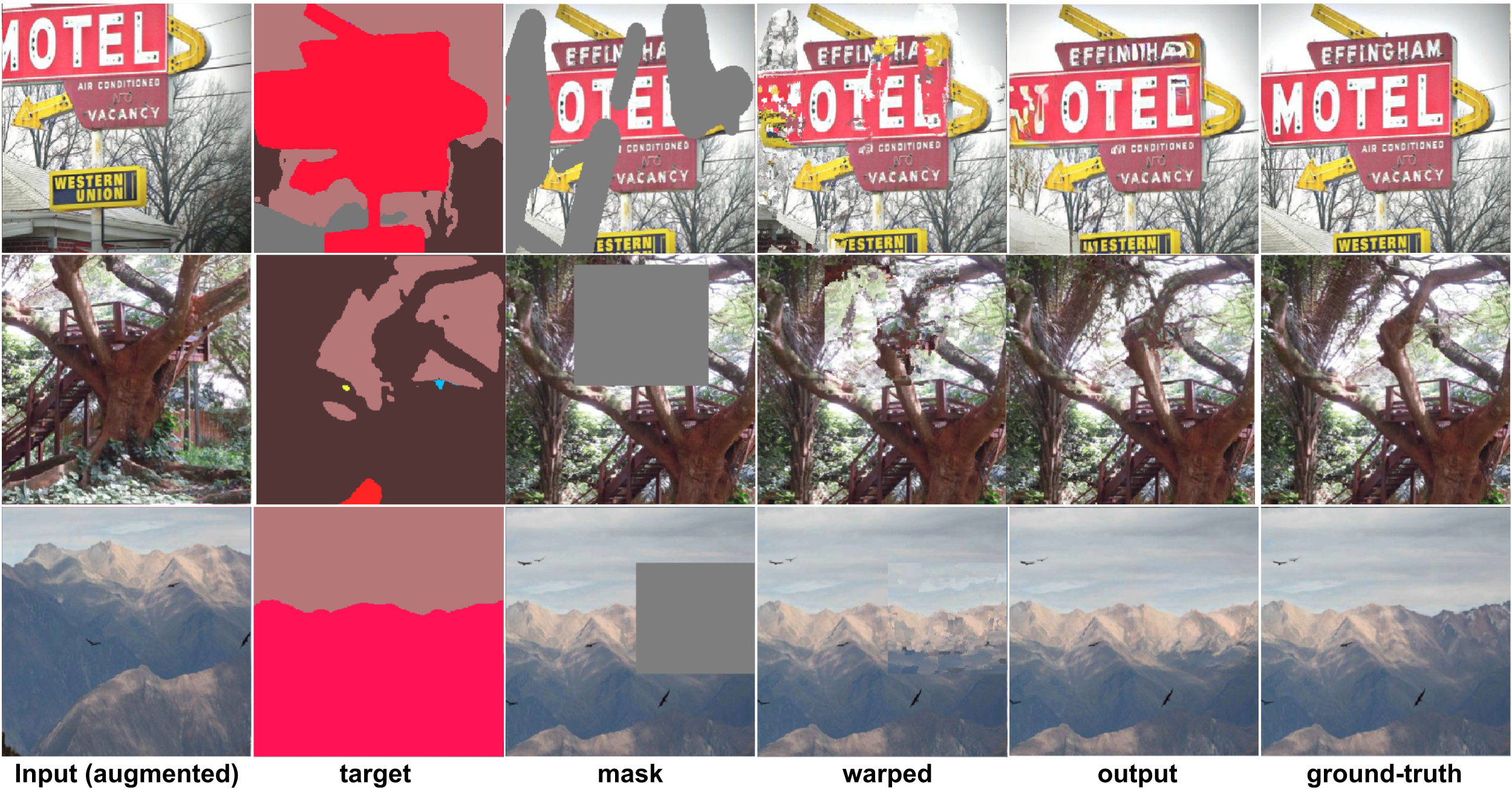}
	\caption{
		\textbf{Visualization of the inputs, outputs and the warped image during the self-supervised training.} Our self-supervised training generates augmented images with new appearance and layout to facilitate generalization to new layout.
	}
	\label{fig:self_supervised_training}
\end{figure}

\begin{figure}[t]
	\centering
	\includegraphics[width=1.0\linewidth]{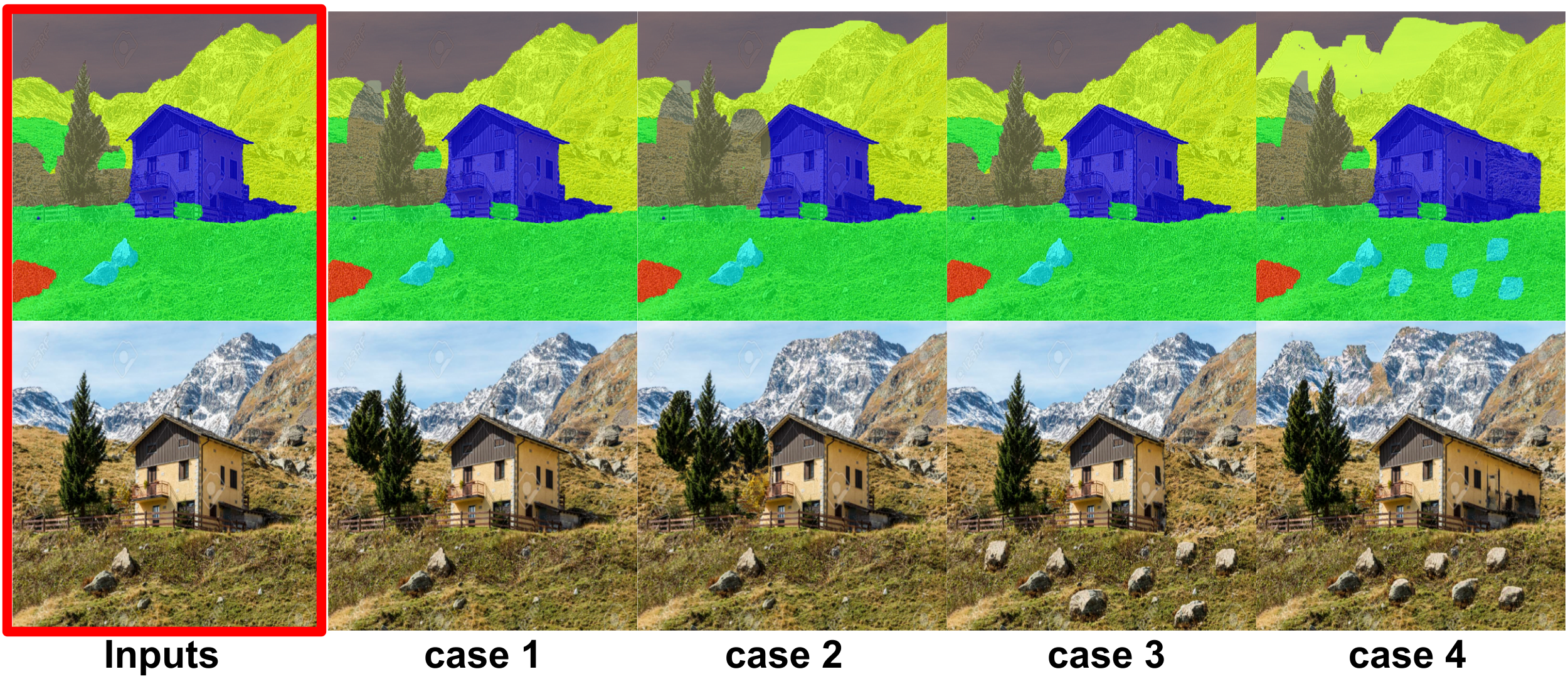}
	\caption{
		\textbf{Visualization of various degrees of manipulation.} Best viewed (e.g. local textures) with zoom-in on screen.
	}
	\label{fig:manipulation}
\end{figure}

To understand why the self-supervised training can generalize to more arbitrary layout change, we provide visualization of the warped image and the final output that are produced during the self-supervised training. As shown in Fig.~\ref{fig:self_supervised_training}, due to the large translation degree that are applied during training, the augmented image and the ground-truth are often different in appearance and layout. 
As results, instead of directly copying the content from the augmented image, our correspondence model can fill in new content inside the mask (e.g. the sign, trunk and the mountain). Fig.~\ref{fig:manipulation} presents the visual examples of our approach under various degrees of layout change, demonstrating the generalization capacity of our model to both small and more significant manipulation requests.

\begin{figure}[]
	\centering
	\includegraphics[width=1\linewidth]{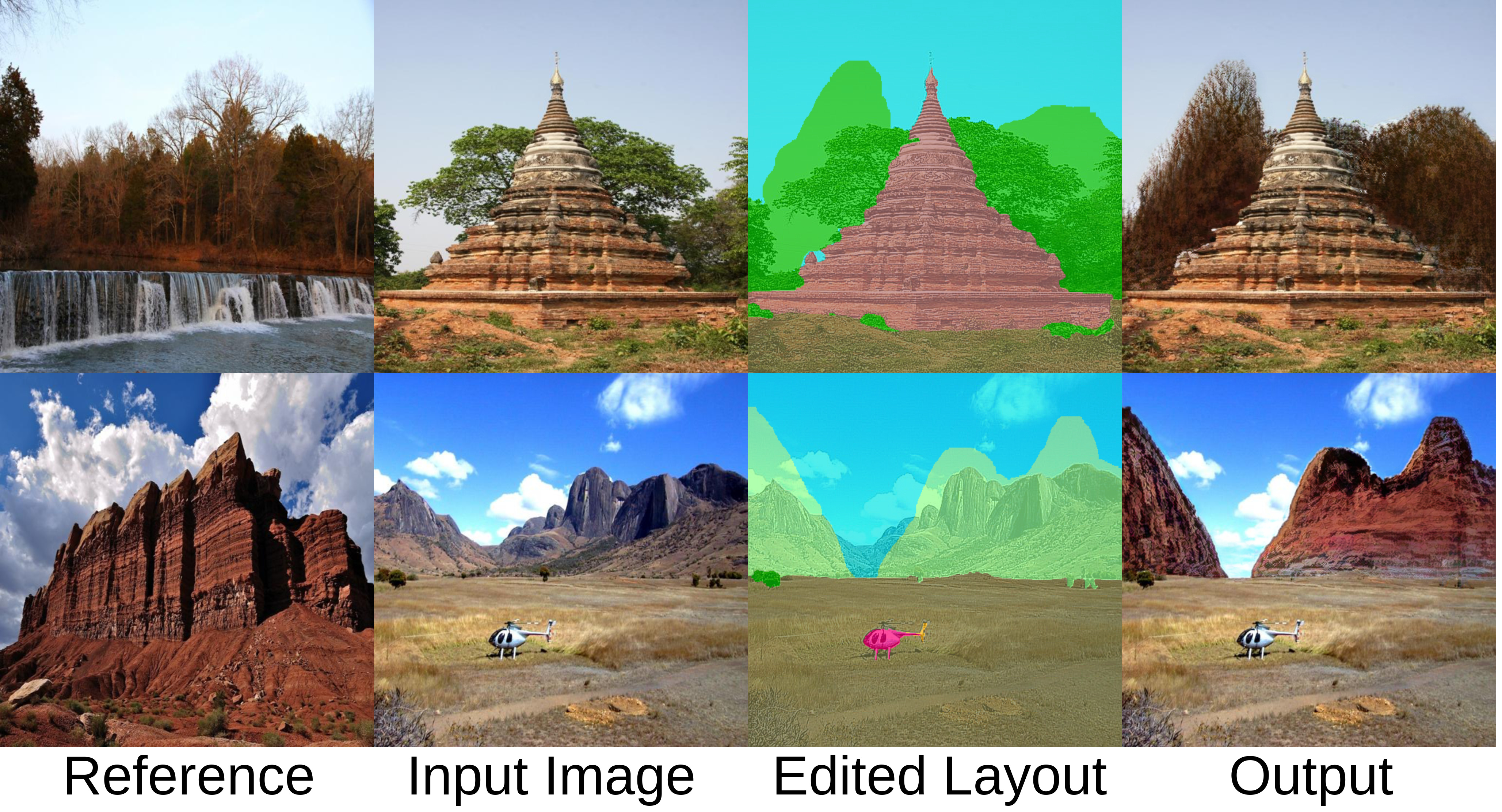}
	\caption{
		\textbf{Example results of our method taking additional reference images to synthesize images with new semantic layouts.} It can also be applied to other multi-image editing tasks such as image compositing and object insertion.
	}
	\label{fig:application}
	\vspace{-0.2in}
\end{figure}

\subsection{Reference-based Layout Manipulation}
\label{sec:application}
\vspace{-0.05in}
Our model is flexible and can also take additional reference images as input for advanced layout manipulation. Some sample results are shown in Fig.~\ref{fig:application}. The results show that our method can seamlessly re-synthesize and compose contents from multiple image sources.

\subsection{Real-world Images Manipulation}
\begin{figure*}[]
	\centering
	\includegraphics[width=0.96\linewidth]{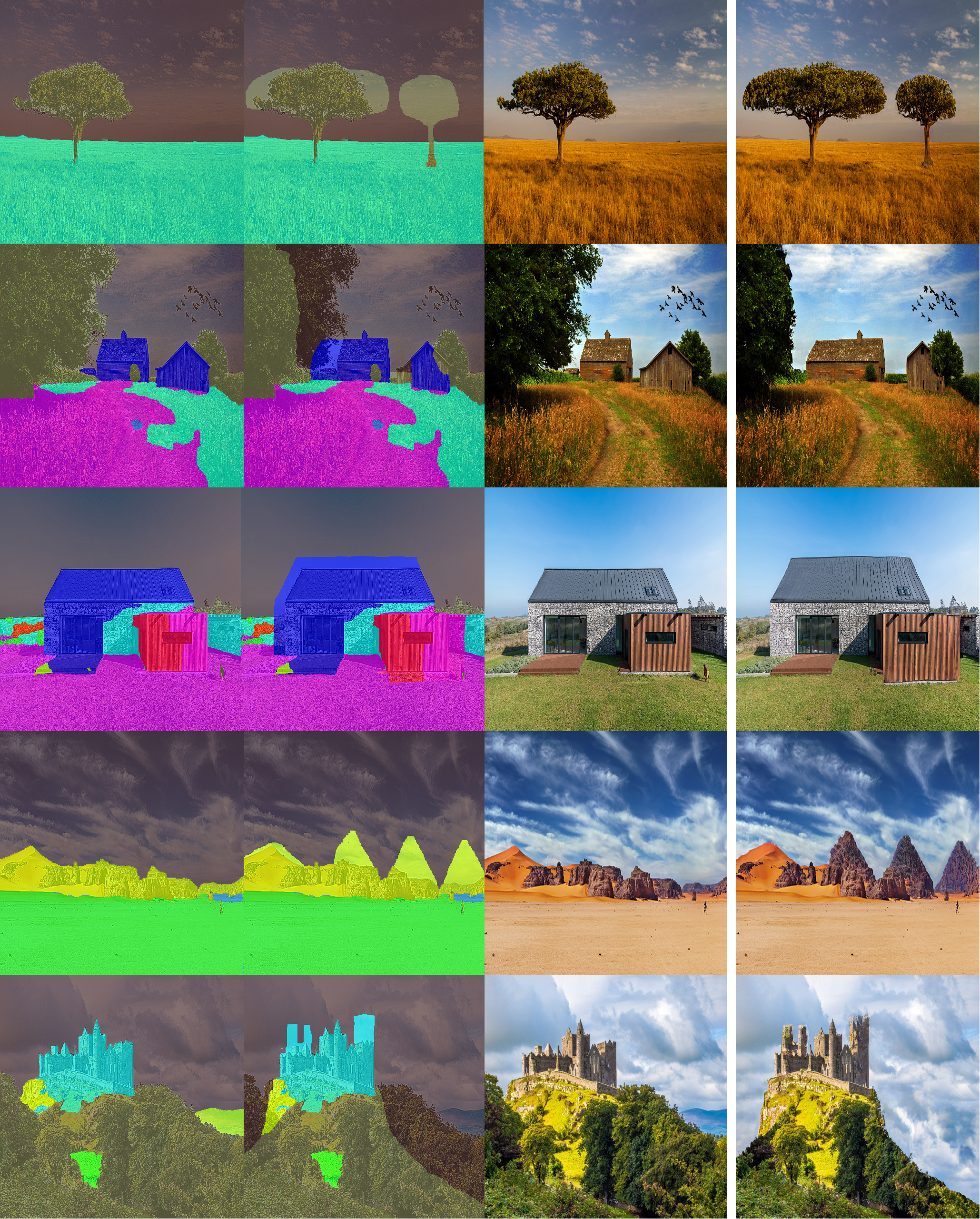}
	\caption{
	\textbf{Qualitative result on the image manipulation task.} From left to right: original layout, edited layout, input images, and edited images. Best viewed (e.g. local textures) with zoom-in on screen.}
	\label{fig:manipulation2}
\end{figure*}

\begin{figure*}[]
	\centering
	\includegraphics[width=1.0\linewidth]{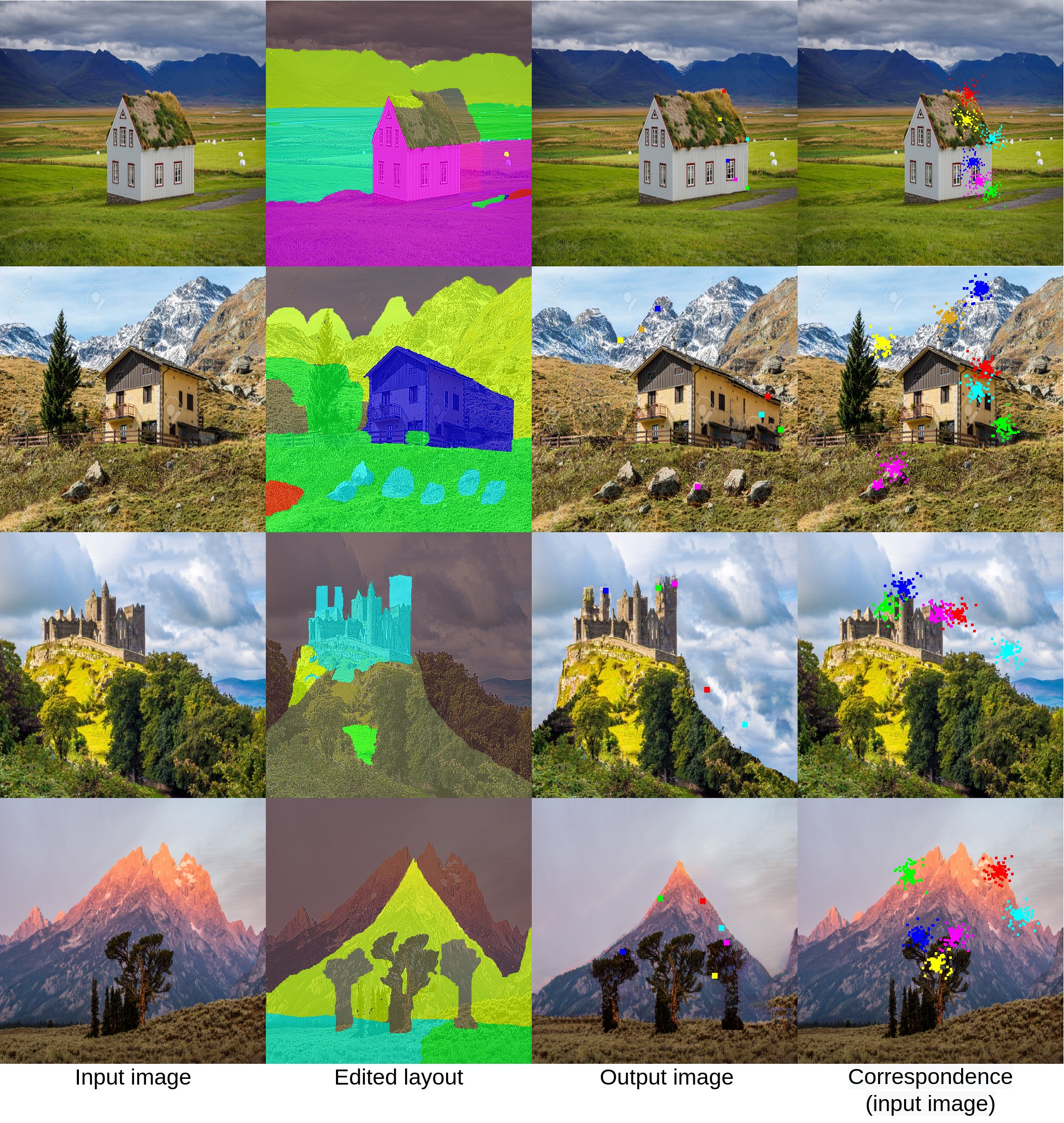}
	\caption{
	\textbf{Correspondence generated by our sparse attention module.} 
	The colored points in 3rd and 4th columns represent the query position and the sampled sparse keys, respectively.}
	\label{fig:correspondence-all}
\end{figure*}

\begin{figure*}[]
	\centering
	\includegraphics[width=0.96\linewidth]{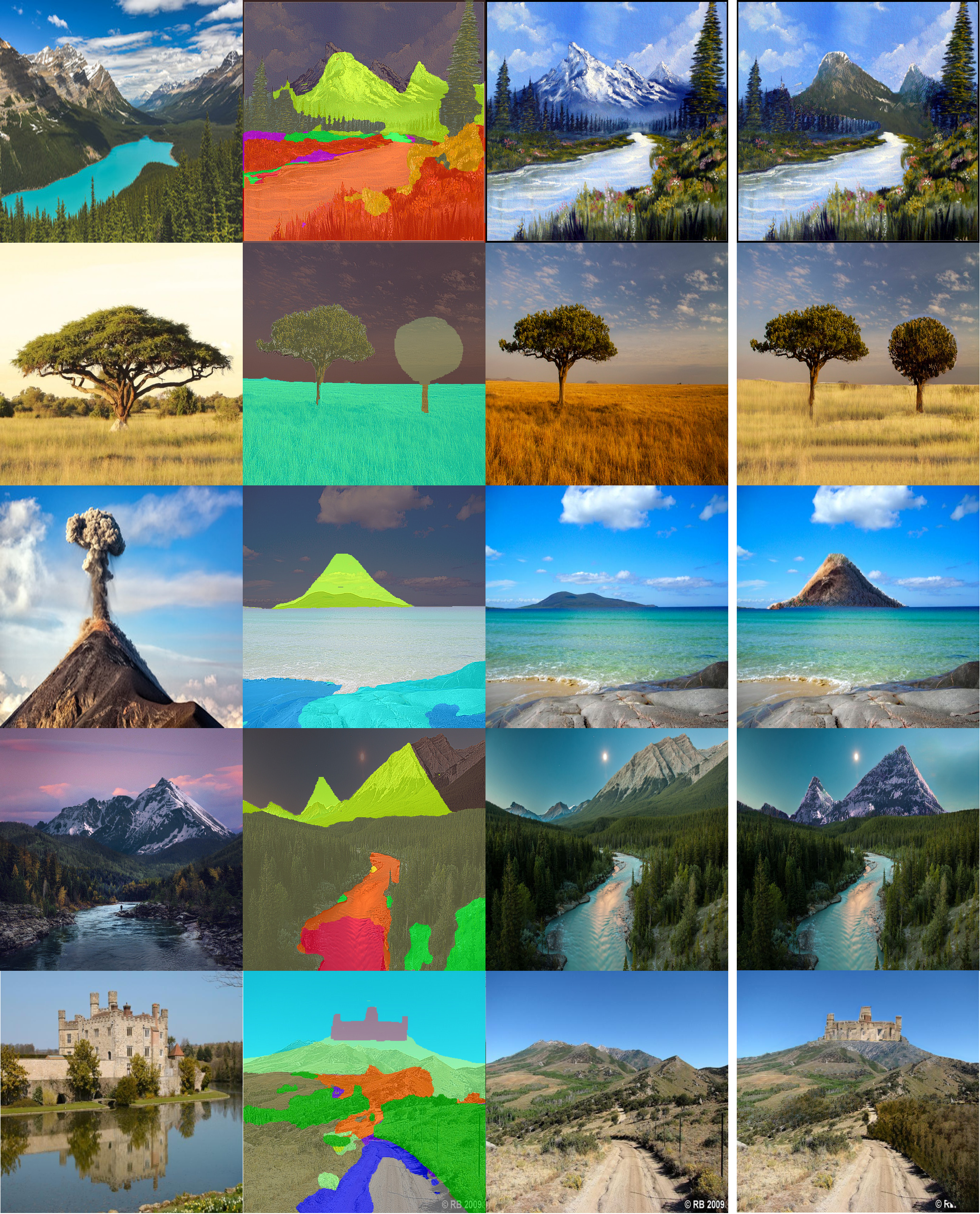}
	\caption{
		\textbf{Qualitative result on the reference-based layout manipulation task.} From left to right: original layout, edited layout, input images, edited images. Best viewed (e.g. local textures) with zoom-in on screen.}
	\label{fig:application-all}
\end{figure*}

To understand how well our model can perform on real-world images, we build an interactive demo for semantic layout manipulation. We provide visual manipulation results in Fig.~\ref{fig:manipulation2}. In Fig.~\ref{fig:correspondence-all}, we provide more qualitative results of the correspondence generated by our high-resolution sparse attention module where the colored points in the 3rd and 4th columns of the figure represent the query position and the sampled sparse keys, respectively. Furthermore, Fig.~\ref{fig:application-all} presents more high-quality visual examples for the reference-based layout manipulation task.

\label{sec:reference}
\section{Conclusion and Future Work}
\vspace{-0.05in}
In this paper, we investigate a challenging semantic layout manipulation task and present a high-resolution sparse attention module that can transfer high-resolution visual details from input images. We leverage our proposed key index sampling step and sparse attentive warping step to enable efficient computation. Based on the trained attention module, we also present a semantic guided generator architecture consisting of a semantic encoder and a two-stage decoder for coarse-to-fine synthesis. We perform extensive experiments on the ADE20k and Places2 datasets to validate the effectiveness of our method on both semantic-guided inpainting and semantic layout manipulation.
Our current approach applies warping at the highest-resolution, thus a multi-scale design would benefit the generated warping. Moreover, the current key index sampling procedure is non-differentiable. Better sampling strategies should be investigated to facilitate better sparse attention computation. Aiming at improving efficiency and performance of our current approach, in the future, we plan to explore a multi-scale correspondence finding design that propagates correspondence from initialized correspondence at a lower resolution. 

\appendices

\section{More Qualitative Comparisons}
\label{sec:visual2}
\begin{figure*}[h]
	\centering
	\includegraphics[width=0.85\linewidth]{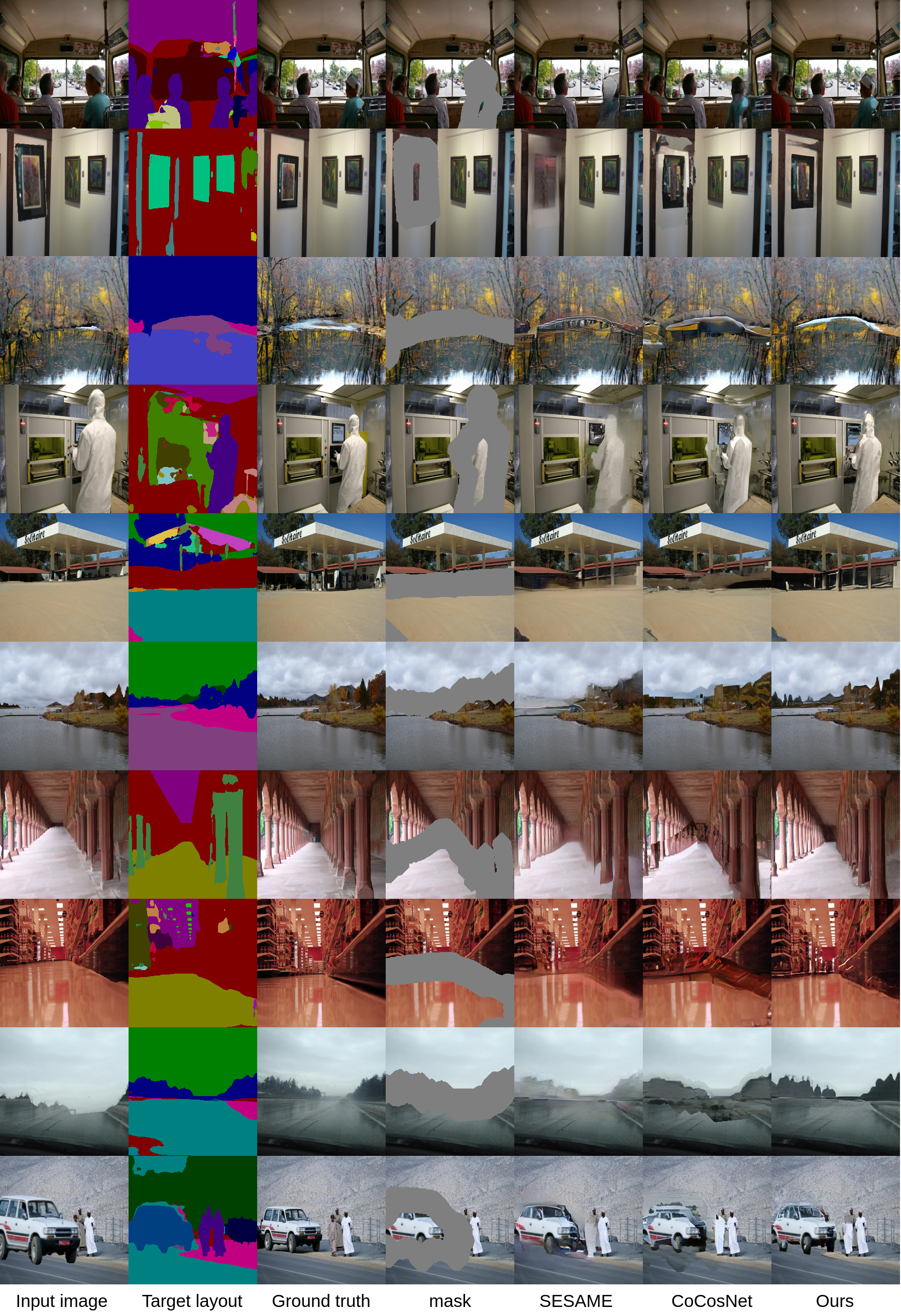}
	\caption{
	\textbf{Qualitative comparisons on the image manipulation task (Places365 dataset).} Best viewed (e.g. local textures) with zoom-in on screen.}
	\label{fig:compare_manipulation}
\end{figure*}

\begin{figure*}[t]
	\centering
	\includegraphics[width=1.0\linewidth]{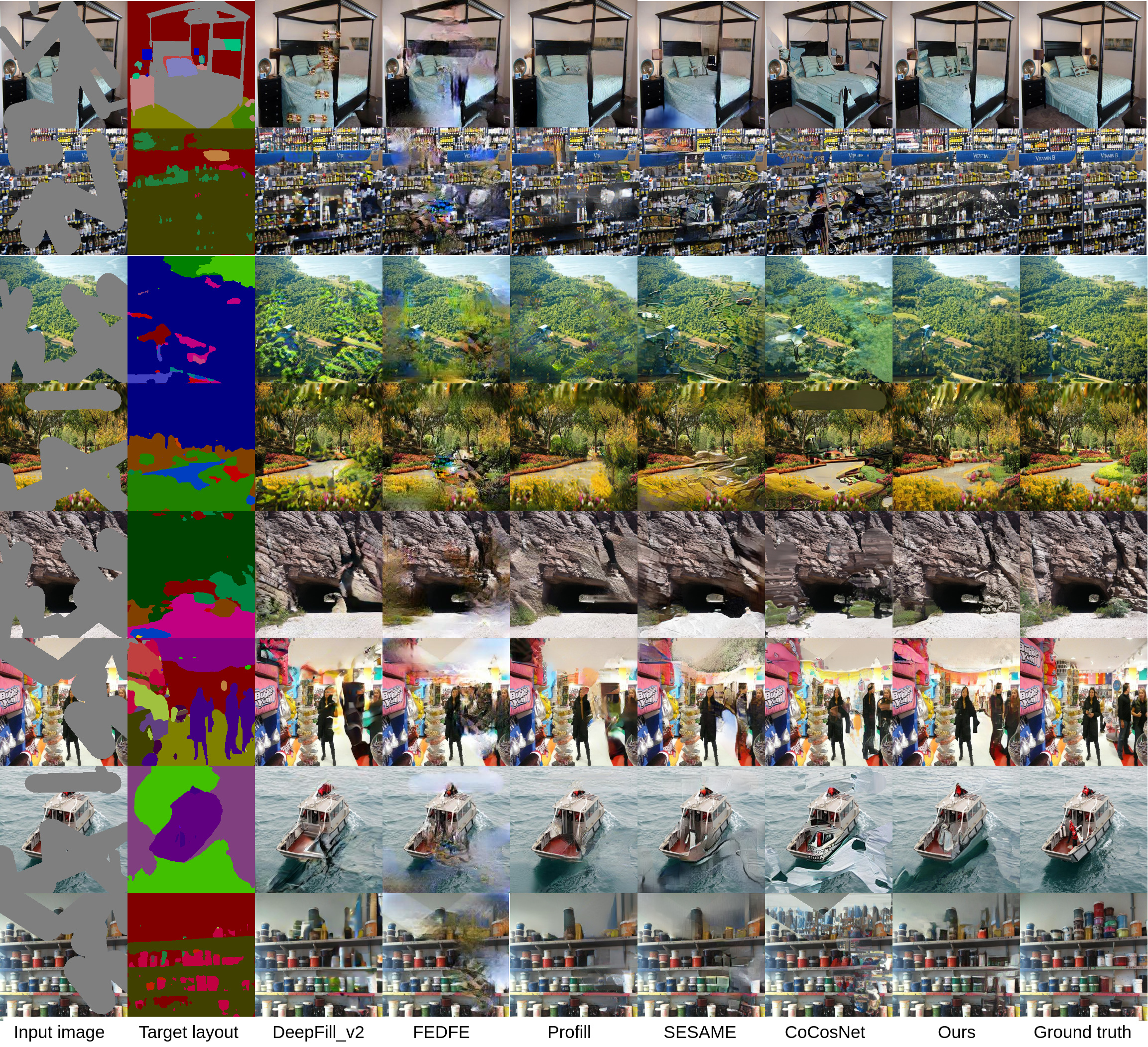}
	\caption{
	\textbf{Qualitative comparisons on the guided reconstruction task (Places365 dataset).} Best viewed (e.g. local textures) with zoom-in on screen.}
	\label{fig:compare_guided_inpainting}
\end{figure*}

\begin{figure*}[t]
	\centering
	\includegraphics[width=1.0\linewidth]{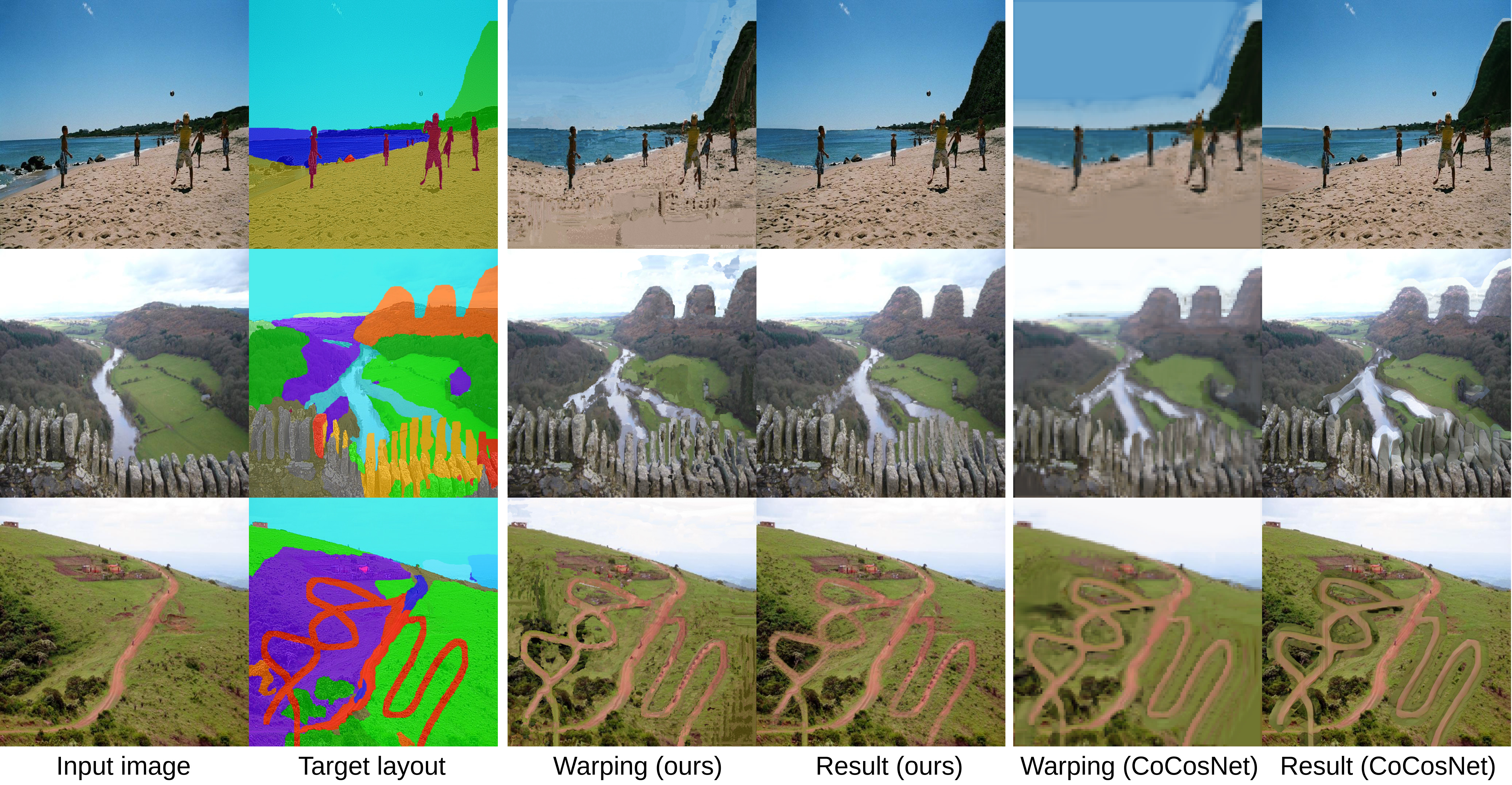}
	\caption{
	\textbf{Visual comparison against CoCosNet. } 
	Our model generate high-resolution warping results to facilitate detail-preserving manipulation. Note that for the results of CoCosNet (the last column), the edited regions are often blurry (e.g., mountain, rock and path for the three rows, respectively).
	Best viewed (e.g. local textures) with zoom-in on screen.}
	\label{fig:compare_real_manipulation}
\end{figure*}

In Fig.~\ref{fig:compare_manipulation} and Fig.~\ref{fig:compare_guided_inpainting}, we provide more qualitative comparisons on the image manipulation and the guided reconstruction task on the Place365 dataset, respectively. In Fig.~\ref{fig:compare_real_manipulation}, compare the intermediate warping results of our model in comparison to CoCosNet~\cite{cocosnet}.

\vspace{-2mm}
\section{Pseudo Code of Key Index Sampling}
We provide the pseudo code in Algorithm~\ref{algorithm:supp_sampling} to sketch the key index sampling procedure. We follow the naming of tensor operations in PyTorch, which include tensor slicing and $\tensor()$, $\convd()$, $\view()$, $\aaa()$, $\cat()$, $\unsqueeze()$, $\sumtorch()$. We use $U(0,1)$ to denote a tensor with its elements uniformly sampled from $[0,1]$. We use $I(K,K)\in \mathbb{R}^{K \times K}$ to denote an identity matrix of dimension $K \times K$. 

\begin{algorithm*}[h]
\DontPrintSemicolon
\SetAlgoLined
\SetNoFillComment
\LinesNotNumbered 
 \KwData{$u_x, u_c \in \mathbb{R}^{B \times C \times H\times W}$ where $B$ and $C$ are the batchsize and channel size, respectively}
 \KwData{$N$: the number of iterations}
 \KwData{$M$: the number of particles}
 \KwResult{$K \in \mathbb{R}^{B \times MN\times 2\times H\times W}$ key index sets}
 \tcc{Initialize 3d kernel $w_v, w_h \in \mathbb{R}^{3 \times 1 \times 1 \times 1\times 3}$ for propagation}
 $w_v \gets I(3,3).\view(3,1,1,1,3)$\;
 $w_h \gets I(3,3).\view(3,1,1,3,1)$\;
 \tcc{Initialize the particle $t\in \mathbb{R}^{B \times M \times 2 \times H\times W}$ and the sparse key index set $K\in \mathbb{R}^{B \times M \times 2 \times H\times W}$}
 $t \gets \tensor(B,M,2,H,W)$\;
 $t[:,:,0,:,:] \gets U(0,1) * H $\;
 $t[:,:,1,:,:] \gets U(0,1) * W $\;
 $K \gets t$\;
 \tcc{main iteration}
 \For{i=1...N-1}{
    \tcc{\bf{Initialization Step}}
    $P \gets \tensor(B,M,k,2,H,W)$\;
    $P[:,:,0,:,:,:] \gets t $ \tcp*{initialize the first element of $P$ as $t$}
    $P[:,:,1:k,0,:,:] \gets t[:,:,0,:,:] + U(0,1) * w;$\;
    $P[:,:,1:k,1,:,:] \gets t[:,:,1,:,:] + U(0,1) * w$\;
    $P[:,:,:,0,:,:] \gets \clamp(P[:,:,:,0,:,:], 0, H)$\;
    $P[:,:,:,1,:,:] \gets \clamp(P[:,:,:,1,:,:], 0, W)$\;
    \tcc{\bf{Propagation Step}}
    $P[:,:,:,0,:,:] \gets \convd(P[:,:,:,0,:,:], w_v)$ \tcp*{vertical propagation}
    $P[:,:,:,1,:,:] \gets \convd(P[:,:,:,1,:,:], w_h)$ \tcp*{horizontal propagation}
    \tcc{\bf{Evaluation Step}}
    $t \gets \tensor(B,M,2,H,W)$\tcp*{initialize $t$}
    $u_x^\prime \gets \gridsample(u_x^\prime, P)$\tcp*{warp $u_x$ using the propagated coordinate $P$}
    $s^\prime \gets \sumtorch(u_x^\prime * u_c, \texttt{dim=1})$\tcp*{compute feature matching scores between $u_c$ and $u_x^\prime$}
    $\texttt{idx} \gets \maxtorch(s^\prime, \texttt{dim=1})$\tcp*{select particles from $P$ that has the maximal matching score}
    $t[:,:,0,:,:]\gets \gathertorch(P[:,:,:,0,:,:], \texttt{idx}, \texttt{dim=3})$ \tcp*{select the y coordinate that has the maximal matching score}
    $t[:,:,1,:,:]\gets \gathertorch(P[:,:,:,1,:,:], \texttt{idx}, \texttt{dim=3})$ \tcp*{select the x coordinate that has the maximal matching score}
    \tcc{\bf{Accumulation Step}}
    $K \gets \cat((K, t), \texttt{dim=2})$ \tcp*{concatenate the updated $t$ to $K$}
 }
 \caption{Key Index Sampling.}
 \label{algorithm:supp_sampling}
\end{algorithm*}




\vspace{-2mm}
\section{Regularization for Training the Sparse Attention Module}
We elaborate the regularization that we mentioned in the training scheme from Section 3.2.
Following the notation from the paper, given an image $x$ and a new layout $c$, we denote the low-resolution feature maps at resolution $H/4 \times W/4$ as $f_x$ and $ f_c$ and the high-resolution feature maps at resolution $H \times W$ as $u_x$ and $u_c$. The low-resolution image is denoted as $x^L \in \mathbb{R}^{H/4 \times W/4 \times 3}$.

To impose the regularization on the feature maps, we first downsample $u_x$ and $u_c$ to resolution $H/4 \times W/4$, denoted by $u_x^L$ and $u_c^L$, respectively.
Then, we impose a cycle-consistency constrain on $u_x^L$ and $u_c^L$ such that the low-resolution image $x^L$ should be consistent to the forward-backward warped image of $x^L$ using the feature maps $u_x^L$ and $u_c^L$. Specifically, the forward warped image $r^L_{\text{forward}}$ is generated by warping $x^L$ using feature map $u_x^L$ and $u_c^L$, $r^L_{\text{forward}}(q) = \sum_{p} {b}^L_{q p} x^L(p)$,
where $p$ iterates over all spatial locations of $x^L$, ${b}^L_{q p} = {\gamma e^{s^\prime_{q p}}}/{\sum_{p} {\gamma e^{s^\prime_{q p}}}}$ is the linear weight computed from the inner-product feature similarity $s^\prime_{q p}=\langle u^L_c(q),u^L_x(p) \rangle$ and $\gamma$ is a temperature coefficient. Similarly, the forward-backward warped image $r^L_{\text{cycle}}$ is generated by warping $r^L_{\text{forward}}$ inversely: $r^L_{\text{cycle}}(q) = \sum_{p} {c}^L_{q p} r^L_{\text{forward}}(p),$
where ${c}^L_{q p} = {\gamma e^{s^\prime_{pq}}}/{\sum_{p} {\gamma e^{s^\prime_{pq}}}}$ is the linear weight again computed from the inner-product feature similarity $s^\prime_{q p}$. Next, the cycle regularization $\mathcal{L}_{cycle}$ is defined as $\mathcal{L}_{cycle} = \norm{r^L_{\text{forward}}-r^L_{\text{cycle}}}^2$.

\vspace{-5mm}
\section{Details of the Manipulation Dataset}
The manipulation evaluation set is generated with the following procedure: first, we sampled an image $x$ from the dataset as the ground-truth image. The semantic label map $c$ of $x$ is taken as the edited label map. To generate the synthetic input image, we repeat the following procedure until the success flag is set to \texttt{True} or the maximal iteration is reached:
\begin{enumerate}
    \item we uniformly sample $10$ positions $p$ from label map $c$ and compute the connected components $g(p)$ of the sampled positions by applying floodfill algorithm. The connected component $g(p^{*})$ with the maximal area is selected.
    \item For the selected component $g(p^{*})$, we compute its convex hull, denoted by $convex(p^{*})$. If $area(convex(p^{*}))/area(p^{*})>0.2$, we skip the current iteration
    \item we uniformly sample scaling factor $s$ in $[1.2,1.5]$ and rotation angle $r$ in $[-15^{\circ}, 15^{\circ}]$ and next apply scaling and rotation on component $g$ on its gravity point, resulting $g^\prime$. If $g^\prime$ cannot cover $g$, we skip the current loop. Otherwise, we set the success flag to \texttt{True} and generate a new image $x^\prime$ with its label map $c^\prime$ by applying the same scaling and rotation operation.
\end{enumerate}
If the success flag is \texttt{True}, we consider $(x^\prime, c^\prime)$ and $(x, c)$ as the image-label pairs before and after editing, respectively. Otherwise, we skip the current image.

\section{Details of the Reconsturction Task}
The reconstruction task aims to reconstruct ground truth images from masked images using an semantic label map as guidance. 
As warping modules of our method and that of CoCosNet~\cite{cocosnet} are designed to use pixels both inside and outside the mask, to make fair comparison with other inpainting methods, we modify our warping stage such that pixels inside the mask are explicitly ignored in the attention-based alignment stage. Specifically, we set large negative values to the correlation matrix of CoCosNet~\cite{cocosnet} for the masked pixels to avoid warping pixels from inside the mask. Likewise, in the evaluation stage of our model, we set large negative values to the matching scores when matched pixels are from inside the mask.

\section{Training details and discussion on the Comparative Methods}
The first and the second stage our our local editing model are trained with batch sizes of 64 and 16, respectively. We use Adam~\cite{adam} as the optimizer with a learning rate of 0.0002 to train both models.  

We compare our method with inpainting methods including  Deepfill-v2~\cite{deepfillv2}, Profill~\cite{profill}, Edge-connect~\cite{edgeconnect}, MEDFE~\cite{liu2020rethinking} and semantic layout manipulation methods including SESAME~\cite{sesame} and global layout editing method CoCosNet~\cite{cocosnet}. Before evaluating performance on the Place365 dataset, we apply a segmentation refinement network~\cite{cascadepsp} to post-process the predicted semantic labels. For the inpainting methods, we use the official code and the pretrained models for evaluation. To evaluate SESAME~\cite{sesame} and CoCosNet~\cite{cocosnet} on the ADE20k dataset, we use the provided official code and the pretrained models. For the evaluation on the Place365 dataset, we train SESAME and finetune CoCosNet on the Place365 dataset.

\ifCLASSOPTIONcaptionsoff
  \newpage
\fi

\begin{IEEEbiography}[{\includegraphics[width=1in,height=1.25in,clip,keepaspectratio]{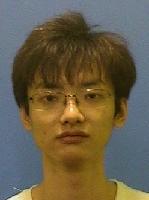}}]{Haitian Zheng}
Haitian Zheng received the B.Sc. and the M.Sc. degrees in electronics engineering and informatics science from the University of Science and Technology of China, under the supervision of Prof. Lu Fang, in 2012 and 2016, respectively. He is currently pursuing the PhD degree with the Computer Science Department, University of Rochester, under the supervision of Prof. Jiebo Luo.
His research interests include image enhancement, generation and manipulation.
\end{IEEEbiography}

\begin{IEEEbiography}[{\includegraphics[width=0.95in,height=1.25in,clip,keepaspectratio]{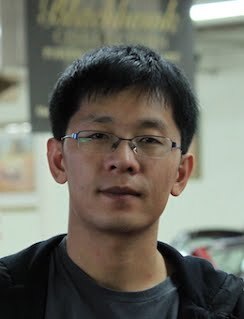}}]{Zhe Lin} is Senior Principal Scientist in Creative Intelligence Lab, Adobe Research. He received his Ph.D. degree in Electrical and Computer Engineering from University of Maryland at College Park in May 2009. Prior to that, he obtained his M.S. degree in Electrical Engineering and Computer Science from Korea Advanced Institute of Science and Technology in August 2004, and B.Eng. degree in Automation from University of Science and Technology of China. He has been a member of Adobe Research since May 2009. His research interests include computer vision, image processing, machine learning, deep learning, artificial intelligence. He has served as a reviewer for many computer vision conferences and journals since 2009, and recently served as an Area Chair for WACV 2018, CVPR 2019, ICCV 2019, CVPR 2020, ECCV 2020, ACM Multimedia 2020.
\end{IEEEbiography}

\begin{IEEEbiography}[{\includegraphics[width=0.95in,height=1.25in,clip,keepaspectratio]{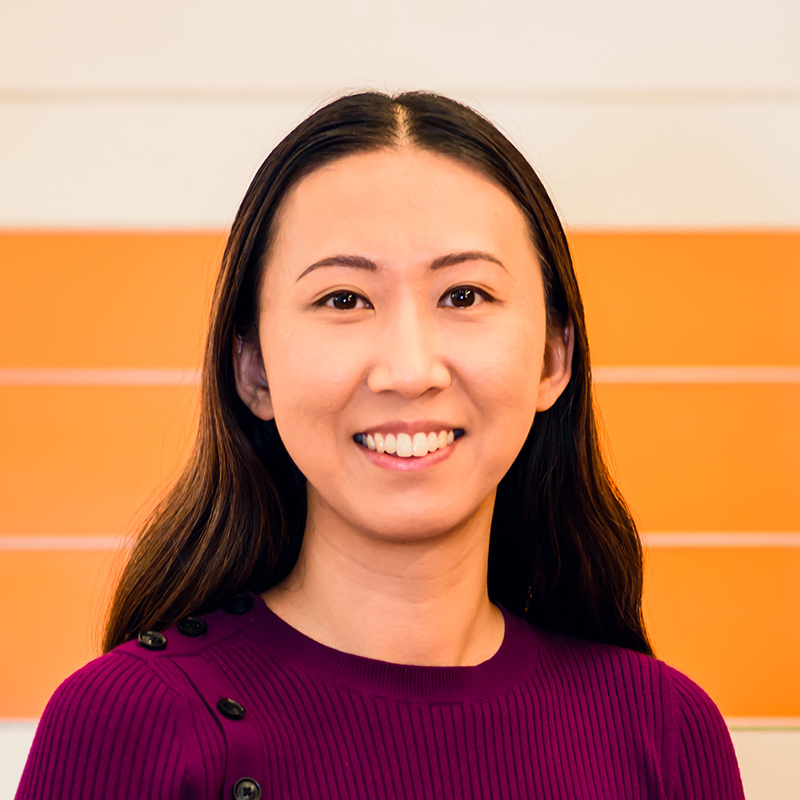}}]{Jingwan Lu} joined Adobe Research in 2014. She is currently leading a team of research scientists and engineers with the vision to disrupt digital imaging and design with big data and generative AI. Her research interests include generative image modeling (GANs, etc.), computational photography, digital human and data-driven artistic content creation. She has over 20 issued patents and over 40 publications in top vision, graphics and machine learning conferences. Some of them attracted lots of public attentions for example, Scribbler, VoCo, StyLit, FaceStyle, Playful Palette. Jingwan served as a program committee member for Siggraph, Siggraph Asia and Eurographics.
\end{IEEEbiography}

\begin{IEEEbiography}[{\includegraphics[width=1in,height=1.25in,clip,keepaspectratio]{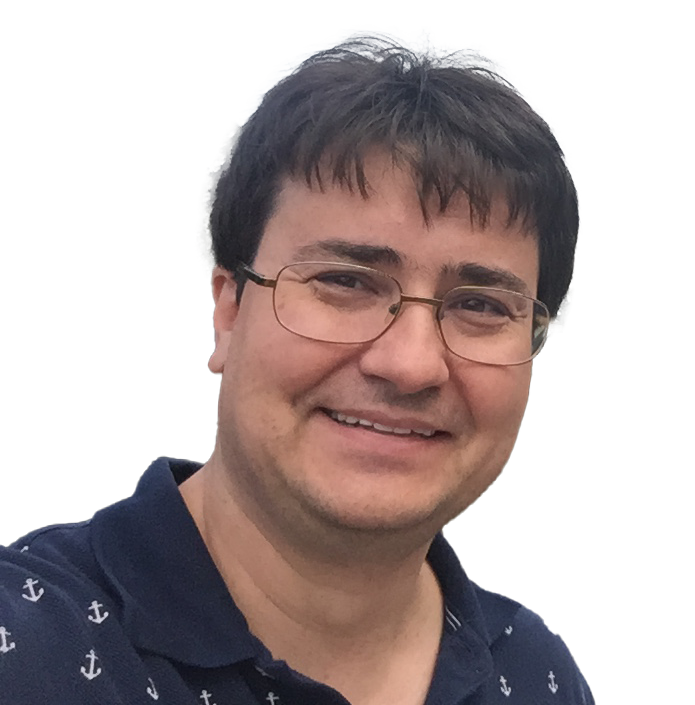}}]{Scott Cohen}
received the B.S. degree in mathematics from Stanford University, Stanford, CA, in 1993, and the B.S., M.S., and Ph.D. degrees in computer science from Stanford University in 1993, 1996, and 1999, respectively. He is currently a Senior Principal Scientist and manager of a Computer Vision research team at Adobe Research, San Jose, CA. His research interests include image segmentation, editing, and understanding such as object attribute prediction, visual grounding, captioning, and search.
\end{IEEEbiography}

\begin{IEEEbiography}[{\includegraphics[width=1in,height=1.25in,clip,keepaspectratio]{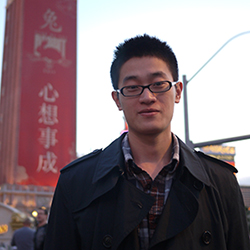}}]{Jianming Zhang}
	is a senior research scientist at Adobe. His research interests include visual saliency, image segmentation, 3D understanding from a single image and image editing. He got his PhD degree in computer science at Boston University in 2016.
\end{IEEEbiography}

\begin{IEEEbiography}[{\includegraphics[width=1in,height=1.25in,clip,keepaspectratio]{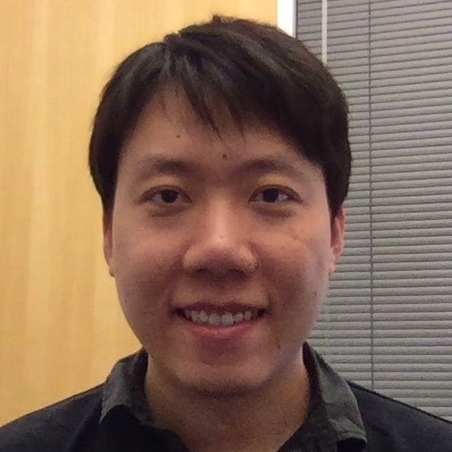}}]{Ning Xu} is a research scientist in Adobe since 2018. He received his PhD degree in Electrical and Computer Engineering from University of Illinois at Urbana-Champaign in 2017, advised by Prof. Thomas Huang. Before joining UIUC, he obtained his B.S. degree in Electrical Engineering from Shanghai Jiao Tong University in 2011. His research interests include language-driven image editing, language-based image synthesis and image/video segmentation and matting.
\end{IEEEbiography}

\begin{IEEEbiography}[{\includegraphics[width=1in,height=1.25in,clip,keepaspectratio]{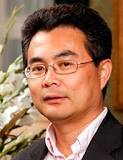}}]{Jiebo Luo}
Jiebo Luo (S93, M96, SM99, F09) is a Professor of Computer Science at the University of Rochester since 2011 after a prolific career of over 15 years with Kodak Research. He has authored over 500 technical papers and holds over 90 U.S. patents. His research interests include computer vision, NLP, machine learning, data mining, computational social science and digital health. He has served as a Program Co-Chair of the ACM Multimedia 2010, IEEE CVPR 2012, ACM ICMR 2016, and IEEE ICIP 2017, and a General Co-Chair of ACM Multimedia 2018. He has served on the Editorial Boards of the IEEE
TRANSACTIONS ON PATTERN ANALYSIS AND MACHINE INTELLIGENCE, IEEE TRANSACTIONS ON MULTIMEDIA, IEEE TRANSACTIONS ON CIRCUITS AND SYSTEMS FOR VIDEO TECHNOLOGY,  IEEE TRANSACTIONS ON BIG DATA,  Pattern Recognition, Machine Vision and Applications, and ACM Transactions on Intelligent Systems and Technology. He is the Editor-in-Chief of the IEEE TRANSACTIONS ON MULTIMEDIA for 2020-2022. Professor Luo is also a Fellow of ACM, AAAI, SPIE and IAPR.
\end{IEEEbiography}




\end{document}